\documentclass[11pt]{article}
\usepackage{makecell}
\usepackage[]{acl}

\usepackage{times}
\usepackage{latexsym}
\usepackage{color}
\usepackage[T1]{fontenc}
\usepackage{graphicx}
\usepackage{subfigure}
\usepackage{multirow}
\usepackage[utf8]{inputenc}
\usepackage{microtype}
\usepackage{float}
\usepackage{adjustbox}
\usepackage{multicol}
\usepackage{placeins}
\usepackage{setspace}
\usepackage{booktabs}

\title{Text Style Transfer Back-Translation}

\author{
  Daimeng Wei\textsuperscript{\rm}\footnotemark[1],
  Zhanglin Wu\textsuperscript{\rm}\footnotemark[1],
  Hengchao Shang\textsuperscript{\rm},
  Zongyao Li\textsuperscript{\rm},\\
  \bf{Minghan Wang\textsuperscript{\rm},}
  \bf{Jiaxin Guo\textsuperscript{\rm},}
   \bf{Xiaoyu Chen\textsuperscript{\rm},} 
  \bf{Zhengzhe Yu\textsuperscript{\rm},} 
  \bf{Hao Yang\textsuperscript{\rm}}\\
  \textsuperscript{\rm}Huawei Translation Service Center, Beijing, China\\
  \tt \{weidaimeng,wuzhanglin2,shanghengchao,lizongyao,wangminghan,\\
  \tt guojiaxin1,chenxiaoyu35,yuzhengzhe,yanghao30\}@huawei.com \\
  }
\begin{document}
\maketitle
\renewcommand{\thefootnote}{\fnsymbol{footnote}}
\footnotetext[1]{These authors contributed equally to this work.}
\renewcommand{\thefootnote}{\arabic{footnote}}
\begin{abstract}
Back Translation (BT) is widely used in the field of machine translation, as it has been proved effective for enhancing translation quality. However, BT mainly improves the translation of inputs that share a similar style (to be more specific, translation-like inputs), since the source side of BT data is machine-translated. For natural inputs, BT brings only slight improvements and sometimes even adverse effects. To address this issue, we propose Text Style Transfer Back Translation (TST BT), which uses a style transfer model to modify the source side of BT data. By making the style of source-side text more natural, we aim to improve the translation of natural inputs. Our experiments on various language pairs, including both high-resource and low-resource ones, demonstrate that TST BT significantly improves translation performance against popular BT benchmarks. In addition, TST BT is proved to be effective in domain adaptation so this strategy can be regarded as a general data augmentation method. Our training code and text style transfer model are open-sourced.\footnote{\url{https://github.com/FrxxzHL/ssebt}}

%But we find that due to the data form of BT, making the input style close to human-translated (HT) content can achieve greater enhancement benefits, and when the style of the input content is more natural, the improvement effect of BT is even smaller, and sometimes there may have a negative effect. We believe that the main reason why HT-style input content benefits more from BT enhancement is that this input is closer to the original text of BT pseudo-corpus, and thus gets better enhancement. Inspired by this, we propose Text Style Transfer BT. This method uses text style transfer to make the original text of the pseudo-corpus more natural, so that BT can greatly improve the nature input. Our experiments on various language pairs (including both high-resource and low-resource ones) demonstrate Style Transfer significantly and robustly improves MT model performance against BT baselines. In addition, Style Transfer is effective in domain adaptation so this strategy can be regarded as a generalized data augmentation method. 

\end{abstract}

\section{Introduction}

Works in neural machine translation (NMT) \cite{sutskever2014sequence,bahdanau2016neural,wu2016googles,vaswani2017attention} greatly improve translation quality. However, current methods generally require large amount of bilingual training data, which is a challenging and sometimes impossible task. As obtaining monolingual data is much easier, researchers have long exploited methods to enhance model performances using monolingual data, for example, language model fusion for phrase-based \cite{brants-etal-2007-large,koehn_2009} and neural machine translation \cite{gulcehre2015using,GULCEHRE2017137}, back translation \cite{sennrich-etal-2016-improving}, and dual learning \cite{cheng-etal-2016-semi,xia2016dual,xia2017dual}. The combination of such monolingual methods can further improve model performances. 

\begin{figure}
\centering
\setlength{\belowcaptionskip}{-0.4cm}
\includegraphics[height=3.5cm,width=8.0cm]{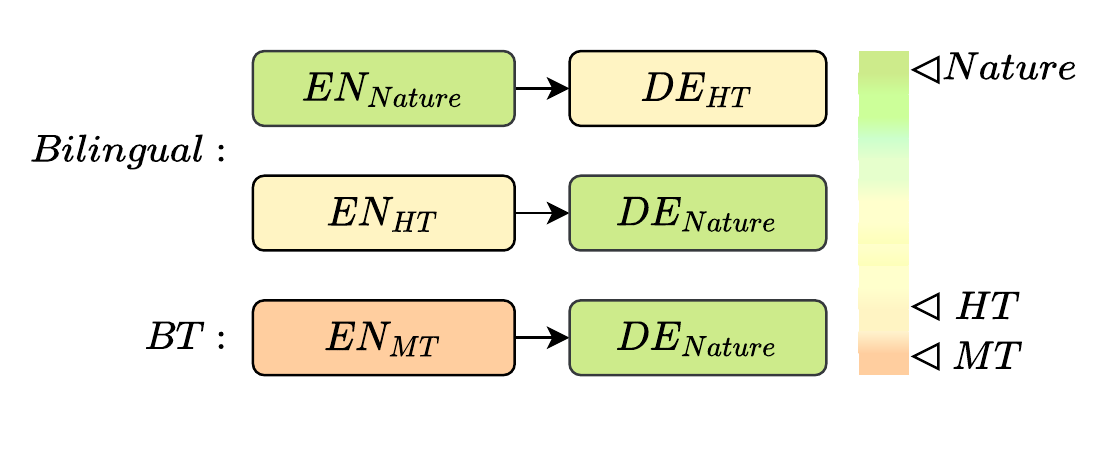}
\caption{Bilingual and BT data used for English $\rightarrow$ German training. $Nature$ indicates data generated by native speakers; $HT$ indicates data generated by human translators from another language, and $MT$ indicates machine translation results. $MT$ and $HT$ styles are close, but far from $Nature$.}
\label{figure:DataTrain}
\end{figure}

Back Translation (BT), a data augmentation method to generate synthetic parallel data by translating content from target language back to source language, is widely used in the field of machine translation. BT has many variants \cite{sennrich-etal-2016-improving,edunov2018understanding,caswell-etal-2019-tagged} and each has own merits. 
% as it is proved effective for performance enhancement

In terms of text style, models that use BT are usually trained on three types of data. Real parallel data constitutes the first two types: natural source with human-translated target ($Nature\rightarrow HT$) or human-translated source with natural target ($HT\rightarrow Nature$). Back translation data constitutes the third type: machine-translated source with natural target ($MT\rightarrow Nature$), as shown in Figure \ref{figure:DataTrain}.

 Inspired by \citet{van-der-werff-etal-2022-automatic}, who find that a classifier can distinguish $MT$ data from $HT$ data, we train a similar classifier to classify $Nature$ and $MT$ data and find that a high percentage of original text is marked as $Nature$ by the classifier. However, the percentage of $Nature$ content is low in human-translated data and even lower in machine-translated data. In general, human and machine translated data are similar, but far different from original text.
 
 We find that when the input style is close to $Nature$, the output is biased towards $HT$; and when the input style is closed to $HT$, the output is biased towards $Nature$ (for details, see Section \ref{sec:6.1}). Since the input used to generate BT data is $Nature$, the output is close to $HT$. So BT mainly improves the translation of translation-like inputs. For natural inputs, BT brings only slight improvements and sometimes even adverse effects. However, in practical use, most inputs sent to NMT models are natural language written by native speakers, rather than translation-like content. 
 %so it is crucial to consider improving translation of such inputs.
 % 此处可以说一下为何一定要解决这个问题：
%因为实际的翻译引擎通常都会接受native speaker 的nature的输入--wmh --done xiaoyu
 
 % We also find (details see sec x.x) that according to the style characteristics of the training data, when the input is close to $Nature$, the translation result is biased toward $HT$ , and when the input content is biased towards $HT$, the translation result will be more closer to $Nature$. Therefore, the style of the source text of the BT pseudo-corpus is more inclined to $HT$. 
 
 %While, the addition of a large amount of $MT \rightarrow Nature$ BT data makes the $MT$ or $HT$ inputs obtain the greatest BT benefit, the $Nature$ inputs benefits little or may bring negative effects.

\begin{table}
\centering
\setlength{\belowcaptionskip}{-0.4cm}
\setlength{\tabcolsep}{1.0mm}{
\begin{tabular}{cc|c|c|c}
\hline
Metrics & Method &Original &Reverse & All\\
\hline
\multirow{2}*{BLEU} & Bitext & \bf{46.3} & 34.9 & 42.2 \\
\cline{2-5}
& BT &41.8 & \bf{42.6} & \bf{42.7} \\
\hline
\hline
\multirow{2}*{COMET} & Bitext & \bf{58.7} & 64.9 & \bf{61.8} \\
\cline{2-5}
& BT & 53.5 & \bf{69.8} & 61.6 \\
\hline
\end{tabular}}
\caption{English$\rightarrow$German BLEU and COMET scores for models trained on WMT 2018 bitext (Bitext) and 24M BT data, measured on WMT 2018 $EN_{Nature} \rightarrow DE_{HT}$ (Original) and $EN_{HT} \rightarrow DE_{Nature}$ (Reverse) test sets.}
\label{tab:01}
\end{table}

We use one original test set ($Nature\rightarrow HT$) and one reverse test set ($HT\rightarrow Nature$) to measure BT performance respectively. As shown in Table \ref{tab:01}, BLEU \cite{post2018clarity} and COMET \cite{rei-etal-2020-comet} scores increase on the reserve test set but decrease on the original test set after BT. 

%The results in table x well reflect this characteristic of BT. Both Bleu \cite{post2018clarity} and COMET \cite{rei-etal-2020-comet} scores on the $HT\rightarrow Nature$ test set increase after applying BT. However, both metrics drop on the $Nature\rightarrow HT$ test set. 

Based on the finding, this paper aims to explore a method to enhance translation of $Nature$ input on basis of BT, while maintaining its effectiveness in translating translation-like content. Since BT connects translation-like input with $Nature$ target, we assume that if we could connect $Nature$ input with $Nature$ target, translation of $Nature$ input could be further enhanced. 

% How to make BT play a greater role in promoting the translation result of $Nature$ input is the focus of this research. The existing BT data cloud connect the $MT$ or $HT$ input with the monolingual target. If we also link the $Nature$ input with the monolingual target, can the translation quality of the $Nature$ input be improved? In other words, can we achieve this goal by creating $Nature\rightarrow Nature$ style BT data without affecting the original advantages of BT.

%将%
Therefore, we propose Text Style Transfer Back Translation (TST BT), aiming to turn $MT\rightarrow Nature$ data into $Nature\rightarrow Nature$ data to enhance the translation of $Nature$ input. However, transferring translation-like text to a natural style is a zero-shot issue, because we can hardly obtain parallel data with the same meaning but different styles ($MT$ and $Nature$).
% 说明zero-shot问题：因为我们很难得到同样意思的机器翻译的结果和native speaker的结果的平行数据--wmh --xiaoyu done
We propose two unsupervised methods. Our experiments on high-resource and low-resource language pairs demonstrate that TST BT can significantly enhance translation of $Nature$ input on basis of BT variants while brings no adverse effect on $HT$ inputs. We also find that TST BT is effective in domain adaptation, demonstrating generalizability of our method.

% To this end we think of the method of using Text Style Transfer (TST). We aim at changing the style of source-side text generated by BT and construct $Nature\rightarrow Nature$ data. However, transferring the style of machine translation to that of natural language is a zero-shot issue. In this paper, we propose two approaches to address it. Our approaches could largely transfer the style of source-side text generated by BT to natural style, and such change significantly leads to performance improvements on $Nature$ input, while, without any side affects on $HT$ input. We call our method as Text Style Transfer Back Translation (TST BT).

% Our experiments on high-resource and low-resource language pairs demonstrate TST BT can bring steady performance improvements on basis of various BT variants. Such improvements mainly comes from the translation of natural inputs. We also find that TST BT is effective in domain adaptation, demonstrating generalizability of our method. 

Our contributions are as follows:
\begin{itemize}
\item We analyze the style of BT text and rationalize its ineffectiveness on $Nature$ input. We herein propose TST BT to solve this issue.
\item TST BT combines Text Style Transfer with BT data to further improve translation of $Nature$ inputs in high and low resource, as well as in-domain and out-of-domain scenarios against various BT baselines.
\item Our experiment results show that TST BT is effective in domain adaptation as well, which further improves model performance on basis of BT augmentation.
% 二三点我觉得可以合并，第一点发现问题，第二点是提出一个通用的解决方案，在高低资源、领域内领域外、还有领域迁移上都有用。

\end{itemize}
\section{Related Work}

%This section presents prior work exploiting source/target-side monolingual data and related work to our synthetic source-enhanced model.
\subsection{Back Translation}
\label{sec:2.1}
Back Translation is first proposed by \citet{bertoldi-federico-2009-domain,bojar-tamchyna-2011-improving} for phrase-based systems, and then applied to neural systems by \citet{sennrich-etal-2016-improving}. 

In general, the standard BT adopts beam search for output generation, so in this paper, we denote it as Beam BT. The following are some BT variants:
\begin{itemize}
\item Sampling BT \cite{edunov2018understanding}: randomly samples translation results based on the probability of each word during decoding, thus largely increases BT data diversity.
\item Noised BT \cite{edunov2018understanding}: adds three types of noise to the one-best hypothesis produced by beam
search. 
\item Tagged BT\cite{caswell-etal-2019-tagged}: adds an extra token to synthetic data to distinguish it from genuine bitext.
\end{itemize}

In our experiment, we use the above four variants as baselines. Other BT variants include Meta BT \cite{pham2021meta}, a cascaded method to supervise synthetic data generation by using bitext information, aiming at generating more usable training data.

\begin{figure}
\centering
\setlength{\belowcaptionskip}{-0.5cm} 
\includegraphics[height=4.3cm,width=8.0cm]{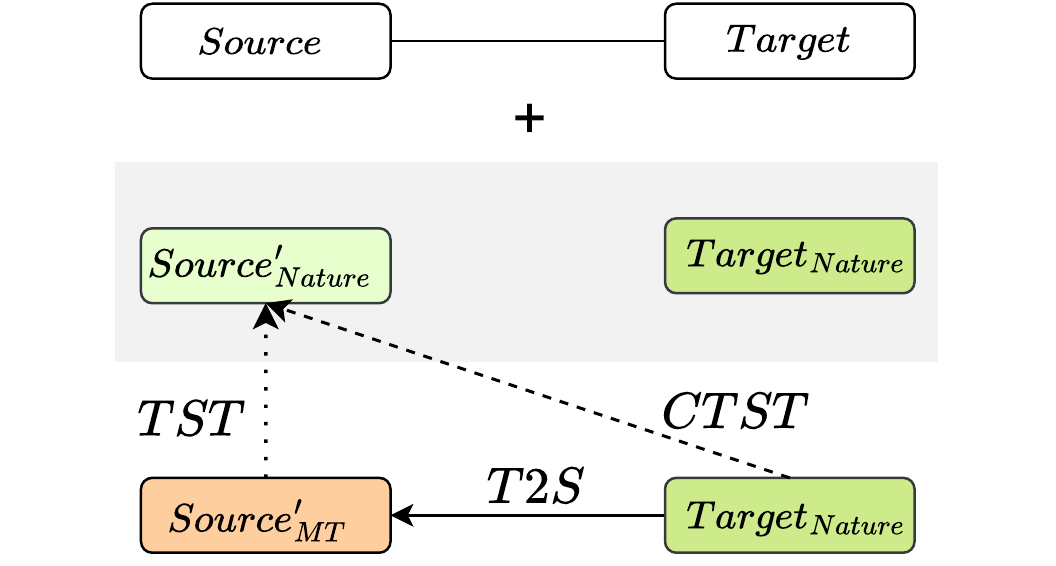}
\caption{Direct and Cascaded methods for TST BT. Source and Target with white color means bilingual data, others mean BT data.}
\label{figure:Method}
\end{figure}

%\citet{edunov2018understanding} propose Sampling BT, believing that diversity of BT data plays a key role in model quality enhancement. They demonstrate the effectiveness of Sampling BT by conducting experiments using large-scale data. Compared with Beam BT, Sampling BT randomly samples translation results based on the probability of each word during decoding, thus largely increases translation diversity. In Sampling BT, authors also propose a diversity version of Beam BT, which adds three types of noise to Beam BT, and name it as Noised BT. The effect of Noised BT is similar to that of Sampling BT. Tagged BT \cite{caswell-etal-2019-tagged} adds an extra token to synthetic data to distinguish from genuine bitext. The authors argue that such strategy also achieves similar results as sampling BT. Meta BT \cite{pham2021meta} is a cascaded method to supervise synthetic data generation by using bitext information, aiming at generating more usable data for model training. In our experiment, we use Beam BT, Noised BT, Sampling BT and Tagged BT as baselines.

\subsection{Unsupervised Text Style Transfer}
Text Style Transfer (TST)  \cite{10.5555/3504035.3504117,10.1162/coli_a_00426}, aiming to control attributes (e.g. politeness) of text, is an important task in the area of natural language generation. Three criteria are used to measure TST: transferred style strength, semantic preservation, and fluency.

As TST training data is difficult to obtain,  unsupervised approaches \cite{dai-etal-2019-style, NEURIPS2018_398475c8, krishna-etal-2020-reformulating, DBLP:conf/ijcai/LuoLZYCSS19} are widely used.  
% xx和xx提出了两种方法和机器翻译并且和风格迁移相关
%xx提出了一种利用分类器+tagging的方法，让Nature的输入，翻译的更Nature，这种方法和我们想要解决的问题很类似，不过这种方法很依赖于双语的数据量，并不能得到稳定的提升。
%xx提出了一种利用大规模译文侧的单语做APE的方法，这种方法也可以认为是一种风格迁移。我们参考了这种训练APE的方法到我们的TST模型训练中。我们的方法和这种方法最大的不同点是，我们发现这种APE的方法在有些情况下并不能提升质量，而结合它带来的风格迁移和BT的特点，则可以稳定的改进BT的提升效果。
Among those, two particular approaches are closely related to machine translation and style transfer. \citet{riley2020translationese} propose using a classifier + tagging approach to make natural input be translated more naturally. This method is similar to the task of our paper, but it has high requirements on bilingual data size and cannot ensure a stable improvement. \citet{freitag2019ape} propose training an Automatic Post-Editing (APE) model with large-scale target-side monolingual data. The APE model can also be considered as a natural style transfer.

We design our TST model by referring to the APE approach. The biggest difference between TST and APE is that APE lacks the ability to improve translation overall quality in some cases, while TST, which combines the advantages of style transfer and back translation, can achieve stable improvements on basis of standard BT.

%\subsection{APE with RTT}
% APE被广泛的研究xx,他们的方法是联合原文和MT结果以及译文来训练APE模型。而我们关注的是只利用单语数据做增强。xx提出了一种的方法
%\noindent A lot of researches \cite{correia-martins-2019-simple,pal-etal-2016-neural,yang-etal-2020-hw} have been done on Automatic Post-Editing (APE). In general, APE leverages three types of data (source sentences, MT results and human translations) to train an APE model. \citet{freitag2019ape} propose the RTT strategy for training an APE model, which incorporates training and inference processes. Differs from other APE models, RTT only focuses on monolingual enhancement. This strategy trains two models on bitext for $X{\rightarrow}Y$ and $Y{\rightarrow}X$, and then generates a round-trip translation $X{\rightarrow}Y{\rightarrow}X^{'}$ to create a noised synthetic dataset $X^{'}=RTT(X)$. Then, an encoder-decoder model is trained to optimize the following objective function:
%\begin{equation}
%1/|D|\sum_{(x,x^{'})\in{D}}{\log{p(x|x^{'})}}
%\end{equation}
%where $D$ is the training data.

%The inference process is as follows: (1) translate source text from language $X$ to language $Y$ with a given translation model, and (2) post-edit the translation outputs to more natural texts with a trained APE model. The training process is rather simple and the post-edited results appear to be more natural and better in quality. 

%Different from \citet{freitag2019ape}, we employ this strategy to augment the BT-generated source-side data. 

\section{Method}
% 我们的方法集中在如何利用风格迁移，产生出风格更接近nature的伪语料原文。为此，我们使用了两种方法：
%- 级联：我们先制造出BT的伪语料，然后对伪语料的原文做迁移。这种方法可以和当前的BT策略解耦合，风格迁移和制造BT数据的Reverse模型可以单独的训练，而且风格迁移训练可以使用其他形式的数据。
%- 直接：我们在制造BT数据的时候，控制生成伪语料原文的风格，尽可能的让生成的伪语料原文偏向于nature。这种方式从风格迁移的角度看，是一种conditional style transfer。它相对于级联模型，在风格迁移的时候，还考虑到了输入的信息，可能对于语义的保留更好。
%不过两种方法都会面临一个问题，无论是第一种的让MT结果的内容迁移到nature风格，还是第二种的直接让Nature的输入翻译成Nature的结果，他们都是零资源的问题。
%1. 级联
%我们先训练两个翻译模型T2S和S2T，我们利用原文单语，采用RTT的方法，这种方法在无监督的风格迁移中广泛的使用xx。如公式x，我们最终制造出风格迁移的训练数据{xx,xx}，并利用它来训练风格迁移器模型。具体的过程可以参见图x (a)。
%直接
%为了让nature的输入翻译的更nature，并且尽可能不像xx一样，受到双语量的影响。我们受启发于xx，采用了两步走的训练策略，如图x(b)所示。
%我们先分别利用原文和译文单语制造出$S_{nature}2T_{mt}$的$S_{mt}2T_{nature}$的数据；然后我们再单独使用使用$S_{nature}2T_{mt}$训练模型；接着我们使用$S_{mt}2T_{nature}$在前面训练的模型上做continue training，不过在这个时候我们冻住encoder的所有参数，只训练cross attention和decoder的参数。
%这种两步走的方法来训练反向模型，试图让模型具备$T_{nature}2S_{nature}$的能力，从而在推理的时候，直接使用这个模型来制造BT数据。
We propose cascaded and direct approaches (see Figure \ref{figure:Method}) to transfer the style of source-side BT data.

%\noindent \textbf{Cascade Mode}: Generate standard BT data and perform style transfer on the source-side text. This approach decouples style transfer with standard BT. The reverse translation model for BT and the style transfer model can be trained independently, and other monolingual data can be used for training the transfer model.

%\noindent \textbf{Direct Mode}: Control the style of output when generating BT data to make the source-side text biased towards Nature. It can be regarded as a kind of Conditional Text Style Transfer (CTST). Comparing with the cascaded approach, it considers the input context during style transfer so the meaning of the input can be better maintained. 

\begin{figure*}
\centering 
\setlength{\belowcaptionskip}{-0.3cm}
\subfigure[]{
\begin{minipage}{7.0cm}
\centering 
\includegraphics[height=4.7cm,width=6.8cm]{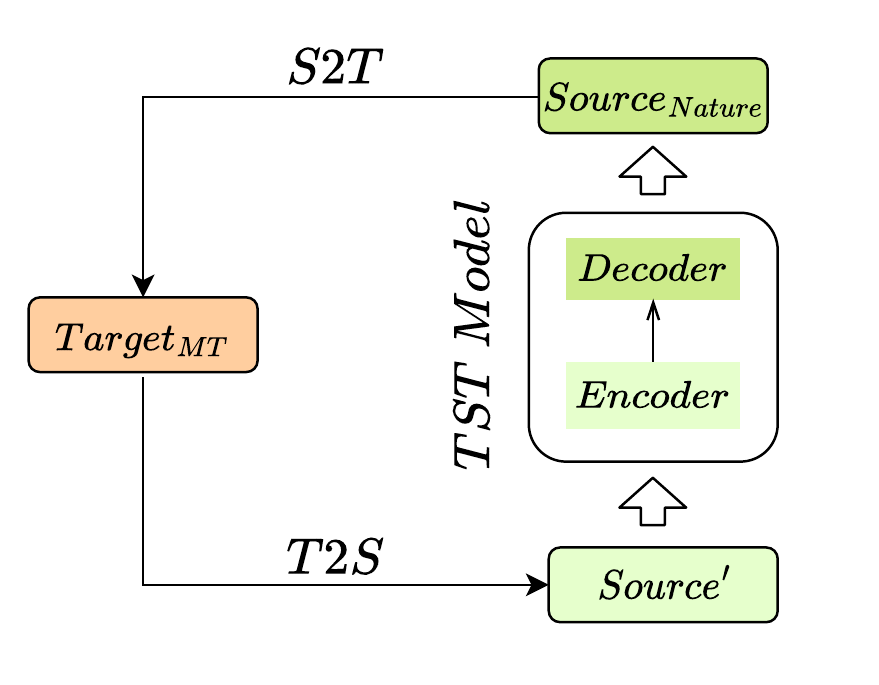}
\vspace{-0.08cm}
\label{fig:RTT}
\end{minipage}
}
\subfigure[]{
\begin{minipage}{8.0cm}
\centering  
\includegraphics[height=4.7cm,width=8.0cm]{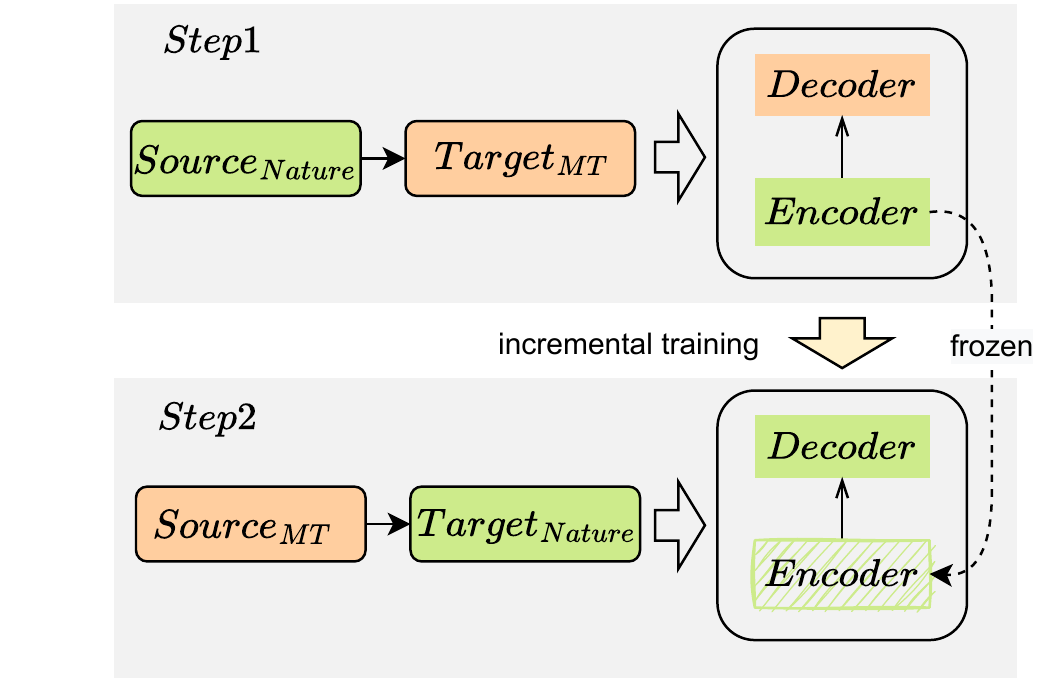}
\vspace{-0.5cm}
\label{fig:Frozen}
\end{minipage}
}
\vspace{-0.3cm}
\caption{Left: TST Model and the process of training data generation. Right: our proposed two-step CTST training scheme.}
\end{figure*}

%However, both of the two approaches, transferring MT to a natural style or directly generating Nature output with Nature input, are zero-shot issues.

\subsection{A Cascaded Approach}
The cascaded approach generates standard BT data first and then modifies the style of the source-side BT data. However, modifying translation-like text to natural text is a zero-shot issue.
To address this, we first train a Source to Target ($S2T$) model and a Target to Source ($T2S$) model. We use the reverse model ($T2S$) to generate BT data $\{Source'_{MT},Target_{Nature}\}$. To generate TST training data, we employ Round Trip Translation (RTT) as shown in formula \ref{equation:01} and Figure \ref{fig:RTT}.
\begin{equation}
Source'=T2S(S2T(Source_{Nature}))
\label{equation:01}
\end{equation}
We use $\{Source', Source_{Nature}\}$ to train the TST model, which uses an encoder-decoder architecture, and apply the model to the source-side BT data $Source'_{MT}$ to get $Nature{\rightarrow}Nature$ data, as shown in formula \ref{eq:02}.
\begin{equation}
Source'_{Nature}=TST(Source')
\label{eq:02}
\end{equation}
The final training data is denoted as:
$$\{(Source,Target),$$
$$(Source'_{Nature},Target_{Nature})\}$$

%For the cascade mode, 
% 我们首先需要解决的是如何制造Translation like->nature风格的平行数据来训练TST。
%we first train two translation models: Source to Target ($S2T$) and Target to Source ($T2S$). Then We employ Round Trip Translation (RTT)
%to generated the TST training data $\{Source', Source_{Nature}\}$, as shown in formula \ref{equation:01} and Figure \ref{fig:RTT}.
%\begin{equation}
%Source'=S2T(T2S(Source_{nature}))
%\label{equation:01}
%\end{equation}
% 需要指出的是，RTT的方法只能得到近似的xx2xx风格的数据。
%After training the TST model, we first generate BT data $\{Source'_{MT},Target_{Nature}\}$ with the reverse model T2S, and then use the TST model to transfer the source side of the synthetic BT corpus, as shown in formula \ref{eq:02}.
%\begin{equation}
%Source'_{Nature}=TST(Source')
%\label{eq:02}
%\end{equation}
%The final training data will be denoted as:
%$$\{(Source,Target),$$
%$$(Source'_{Nature},Target_{Nature})\}$$
\subsection{A Direct Approach}
% 直接将Nature的数据翻译成Nature的结果，也是一个zero-shot问题。
Directly translating $Nature$ data into $Nature$ outputs is also a zero-shot issue \cite{riley2020translationese}. In order to make $Nature$ input be translated more naturally, and avoid the data size limitations mentioned by \citet{riley2020translationese}, we adopt a two-step training strategy, which is inspired by \citet{DBLP:conf/acl/ZhangLL21a}, as shown in Figure \ref{fig:Frozen}.

We first use source and target side monolingual data to generate $Source_{Nature}$ to $Target_{MT}$ and $Source_{MT}$ to $Target_{Nature}$ data respectively. We use only $Source_{Nature}$ to $Target_{MT}$ data to train the translation model and perform incremental training with $Source_{MT}$ to $Target_{Nature}$ data. During incremental training, we freeze all parameters in the encoder so the model only learns decoder parameters. 

By using the two-step strategy, we aim to let the translation model learn how to produce $Nature \rightarrow Nature$ data. %我们认为这种直接的方法是一种Conditional Text Style Transfer (CTST)的方法
We consider this approach as a Conditional Text Style Transfer (CTST) method.

\section{Experimental Setup}
\subsection{Data}
% 我们的主实验在WMT18 ENDE新闻翻译任务，WMT17 ZH-EN新闻翻译任务以及WMT16 EN-RO新闻翻译任务上进行。在WMT18 ENDE新闻翻译任务上，我们使用除ParaCraw语料库以外的5.2M双语数据训练基线模型，并使用226.2M NewsCrawl 2007-2017德语单语数据进行反向翻译。在WMT17 ZH-EN新闻翻译任务上，我们使用19.1M双语数据训练基线模型，在进行反向翻译时，我们使用20.4M News Crawl 2016英语单语数据。在WMT16 EN-RO新闻翻译任务上，我们使用0.6M双语数据训练基线模型，并使用2.2M News Crawl 2015罗马尼亚语单语数据进行反向翻译。
% 在训练TST模型时，我们需要使用源语言单语数据构造训练语料。在WMT18 ENDE和WMT16 EN-RO新闻翻译任务上，我们使用24M NewsCrawl 2007-2017英语单语数据，而在WMT17 ZH-EN新闻翻译任务上，我们使用24M Common Crawl中文单语数据。
% 在域外分析实验中，我们使用WMT18 DE-EN 5.2M新闻双语数据训练基线模型，并基于收集到的2.5M德语医疗单语数据训练TST模型，然后使用收集到的12M英语医疗单语数据进行反向翻译，最后再基于医疗TST模型对反向翻译伪语料进行风格迁移。
Our main experiments are conducted on WMT18 EnDe, WMT17 ZhEn, and WMT16 EnRo news translation data. 
For EnDe, we use 5.2M bilingual data except ParaCraw corpus to train the baseline model, and 226.2M German monolingual data from NewsCrawl 2007-2017 for back translation. For ZhEn, we use 19.1M bilingual data to train the baseline model, and 20.4M English monolingual data from News Crawl 2016 for back translation. For EnRo, we use 0.6M bilingual data to train the baseline model and 2.2M Romanian monolingual data from News Crawl 2015 for back translation.

Training the TST model requires source-side monolingual data. we use 24M English monolingual data from NewsCrawl 2007-2017 for EnDe and EnRo, and 24M Chinese monolingual data for ZhEn.

\subsection{Evaluation}
% 我们使用BLEU，ChrF，COMET以及BLEURT等指标评估NMT模型在测试集上的表现。其中，BLEU和Chrf均采用SacreBLEU工具计算，COMET采用WMT-20-DA模型计算，BLEURT采用BLEURT-20模型计算。我们还基于xlm-roberta-base预训练模型，使用simpletransformers工具训练了一个nature与MT的二分类器，用于后续的分析实验，其训练语料来源于10M英语单语和10M机器翻译英语单语。

We use metrics including BLEU \cite{papineni-etal-2002-bleu}, ChrF \cite{popovic2015chrf}, COMET \cite{rei2020comet} and BLEURT \cite{sellam2020bleurt} to evaluate models performances on test sets. Among them, BLEU and ChrF are calculated using SacreBLEU\footnote{\url{https://github.com/mjpost/sacrebleu}}\cite{post2018clarity}, COMET using wmt20-comet-da\footnote{\url{https://github.com/Unbabel/COMET}}, and BLEURT using BLEURT-20\footnote{\url{https://github.com/google-research/bleurt}}. Based on the xlm-roberta-base\footnote{\url{https://huggingface.co/xlm-roberta-base}} pre-training model, we use simpletransformers\footnote{\url{https://github.com/ThilinaRajapakse/simpletransformers}} to train a binary classifier to classify $Nature$ and $MT$ text for subsequent experiments. The training data includes 10M natural monolingual data and 10M machine-translated monolingual data.

%\subsection{Directional test sets}
%In general, WMT organizers collect monolingual news data of related languages and recruit professional translators to translate the data into corresponding languages. For a language pair, there are generally two test sets: a source-original test set of which the source side is natural language and the target side is translated text, and a target-original test set vice versa. Our experiment measures both test sets so as to better evaluate the impact of our TST model in different scenarios. We denote the test sets similar as \citet{NikolayBogoychev2019DomainTA}: the source-original test sets is denoted as "Original" and target-original test sets as "Reverse".

\begin{table}
\centering
\setlength{\belowcaptionskip}{-0.5cm} {
\begin{tabular}{l|l}
BT type & Example sentence\\
\hline
Beam & Raise the child, love the child.\\
Sampling & Lift the child, love the child.\\
Noised & Raise child \underline{\ \ \ } love child, the.\\
Tagged & <T> Raise the child, love the child.\\ 
\end{tabular}}
\caption{The source text of synthetic corpus for different BT methods}
\label{tab:02}
\end{table}

\subsection{Architecture}
% 我们采用由fairseq工具提供的Tranformer结构，使用FP16加速，并使用源语言和目标语言共享词表的方式来训练NMT模型与TST模型。具体来说，WMT18 ENDE模型，WMT17 ZH-EN模型以及TST模型均采用Transformer-big结构，词表大小为32K，而WMT16 EN-RO模型则采用Transformer-base结构，词表大小为16K。其中，WMT18 ENDE基线模型和TST模型的dropout rate为0.3，其他模型的dropout rate都为0.1。剩余的一些参数设置如下：batch_size大小为4096，学习率为7e-4，warmup步数为4000，label-smoothing为0.1，Adam优化器的β1参数为0.9，β2参数为0.98。对于每个训练任务，我们根据在开发集上测量的困惑来选择最佳模型。

We train our NMT models and TST models with Transformer \cite{vaswani2017attention} and fairseq \cite{ott-etal-2019-fairseq}, and employ FP16 to accelerate training under a joint source and target language vocabulary setting. Specifically, EnDE, ZhEn, and the TST models use the Transformer-big structure with a vocabulary size of 32K, while EnRo models use the Transformer-base structure with a vocabulary size 16K. The dropout rate for EnDe baseline model and TST model is 0.3, and 0.1 for other models. Other settings are as follows: batch size as 4096, learning rate as 7e-4, warmup steps as 4000, label-smoothing as 0.1 \cite{szegedy2015rethinking,pereyra2017regularizing}, Adam ${\beta}1$ as 0.9, and ${\beta}2$ as 0.98 \cite{kingma2017adam}. For each training task, we select the best model according to the perplexities measured on the dev set. 

% For more details about our training parameters, please refer to our open-source code.

\begin{table*}
\centering
\setlength{\belowcaptionskip}{-0.4cm}
\setlength{\tabcolsep}{1.7mm} {
\begin{tabular}{l|ccc|ccc|ccc|ccc}
\hline
 \multirow{2}* & \multicolumn{3}{c|}{BLEU} & \multicolumn{3}{c|}{ChrF} & \multicolumn{3}{c|}{COMET} &
 \multicolumn{3}{c}{BLEURT}\\
 \cline{2-13}
~ &All &O &R &All &O &R &All &O &R &All &O &R \\
\hline
Bitext &32.9 & 35.2 & 28.9 & 60.8 & 62.1 & 59.1 & 54.8 & 50.1 & 59.7 & 73.6 & 71.8 & 75.6 \\
\hline
+Beam BT &32.1 & 28.5 & 36.4 & 59.2 & 55.0 & 65.0 & 45.9 & 28.0 & 65.4 & 71.7 & 66.2 & 77.8 \\
\ \ \ \ +TST$_{Direct}$ & 33.3  & 31.4  & 34.8  & 60.8  & 58.4  & 64.1  & 53.3  & 42.2  & 65.4  & 73.9  & 70.3  & 77.8 \\
\ \ \ \ +TST$_{Cascade}$  &\bf{35.3} & \bf{33.0} & \bf{37.7} & \bf{62.8} & \bf{60.6} & \bf{65.8} & \bf{59.3} & \bf{51.6} & \bf{67.6} & \bf{75.8} & \bf{73.1} & \bf{78.7} \\
\hline
+Sampling BT &\bf{36.0} & \bf{32.7} & \bf{40.2} & \bf{63.0} & 60.2 & \bf{66.9} & 61.7 & 54.5 & 69.5 & 76.9 & 74.2 & 79.7\\
\ \ \ \ +TST  &35.8 & 32.6 & 39.9 & \bf{63.0} & \bf{60.3} & 66.8 & \bf{62.5} & \bf{55.9} & \bf{69.6} & \bf{77.2} & \bf{74.7} & \bf{79.8}\\
\hline
+Noised BT &36.6 & 36.2 & 36.4 & 63.6 & 62.6 & 65.0 & 59.8 & 53.4 & 66.9 & 75.7 & 73.0 & 78.5\\
\ \ \ \ +TST  &\bf{37.0} & \bf{36.5} & \bf{37.1} & \bf{64.1} & \bf{63.1} & \bf{65.5} & \bf{62.3} & \bf{57.1} & \bf{67.9} & \bf{76.5} & \bf{74.4} & \bf{78.9}\\
\hline
+Tagged BT & 37.0 & 36.6 & \bf{36.7} & 63.9 & 63.1 & \bf{64.9} & 61.6 & 56.0 & \bf{67.6} & 76.2 & 73.9 & 78.6\\
\ \ \ \ +TST &\bf{37.4} & \bf{37.4} & 36.5 & \bf{64.3} & \bf{63.8} & \bf{64.9} & \bf{62.2} & \bf{57.2} & \bf{67.6} & \bf{76.6} & \bf{74.4} & \bf{78.9}\\
\hline
\hline
+FT &33.6 & 36.4 & 28.9 & 61.5 & 63.1 & 59.3 & 56.3 & 52.4 & 60.4 & 74.1 & 72.5 & 75.8 \\
\ \ \ \ +Beam BT  &37.3 & 37.6 & 36.1 & 64.3 & 63.8 & 64.9 & 60.4 & 54.7 & 66.4 & 75.6 & 73.3 & 78.0 \\
\ \ \ \ \ \ \ \ +TST &\bf{37.8} & \bf{37.7} & \bf{37.2} & \bf{64.6} & \bf{64.0} & \bf{65.5} & \bf{61.3} & \bf{55.4} & \bf{67.6} & \bf{76.1} & \bf{73.8} & \bf{78.6}\\
\hline
\end{tabular}}
\caption{English$\rightarrow$German models trained on WMT 2018 bitext (Bitext) with four BT variants (Beam, Sampling, Noised and Tagged BT). Their averaged TST results on Original test set (O), Reverse test set (R) and the combined test sets (All) from WMT 2014-2018.}
\label{tab:03}
\end{table*}
\section{Result}
TST can be combined with popular BT strategies. Our strategy can be seen as a universal data argumentation method on basis of BT. To better verify the effectiveness of our method, Beam BT, Sampling BT, Noised BT, and Tagged BT are selected for comparative experiments (see Section \ref{sec:2.1}). 

Table \ref{tab:02} is an example of synthetic source sentences generated by four BT strategies. For Noised BT, noise is added after TST is performed. While for other BT methods, we directly modify the source side of BT data using our TST model.

To prove the effectiveness of TST BT, We perform experiments on high-resource (EnDe and ZhEn) and low-resource (EnRo) languages, as well as domain adaptation.
%\subsection{High Resource}
%We select EnDe and ZhEn for high-resource experiments. The source languages, English and Chinese, belongs to different language families and require two style transfer models for each language. In addition, monolingual and bilingual data sizes for each language pair differ, so we can test the generalizability of TST BT on different data sizes.

\subsection{TST BT for EnDe}
We believe that when we add $Nature$ to $Nature$ BT data, the translation of $Nature$ input can be improved. However, the target side of original test set is human-translated, which could influences the scores measured by over-lapping metrics, such as BLEU and ChrF. For the purpose of fair evaluation, we report multiple metric scores, including BLEU, ChrF, COMET, and BLEURT. The final scores are averaged based on WMT14-18 test sets, as shown in Table \ref{tab:03}. The detail results are shown in Appendix \ref{EnDeDetails}.

All BT methods enhance model performance over baselines. It has greater effect on reverse test sets than original ones. Particularly, all metrics on original test set decline after Beam BT is applied. This result is consistent with our findings that merely adding BT data $MT{\rightarrow}Nature$ deteriorates translation of $Nature$ input. %说明单纯的增加BT数据的确会影响Nature input的翻译质量。-done

We try the two style transfer approaches mentioned above on basis of Beam BT. The result shows that both cascaded and direct approaches bring significant improvements but the cascaded approach is better. So we use the cascaded approach by default in following experiments. 
% Sec x.x details the comparison of the two approaches. 

In general, TST BT mainly brings improvement on original test sets while maintains standard BT's effectiveness on reverse test sets. Although BLEU and ChrF scores are fluctuated, we observe steady increase of COMET and BLEURT scores after TST BT is applied. We observe similar improvements against other BT baselines, with an average improvement of 1.0+ COMET score.
% 除了对Beam BT非常大幅度的提升之外，其他的强BT基线，我们在COMET上也有平均一个点的提升。-done
%In addition to the significant improvement of Beam BT, compared with other strong BT baselines, we also have an average improvement of 1.0+ COMET score.
%这个需要再优化一下，解释一下Bleu和ChrF没有明显改变的原因，引用几篇论文
%However, adding $Nature\rightarrow Nature$ data changes the style of translation. The style change can hardly be measured by over-lapping based methods, but are clearly reflected by COMET and BLEURT scores. We make further analysis in this regard in sec XX.

According to the experiment results, TST is a supplement to BT that further enhances the effectiveness of BT.
\begin{table*}
\centering
\begin{tabular}{l|ccc|ccc|ccc|ccc}
\hline
 \multirow{2}* & \multicolumn{3}{c|}{BLEU} & \multicolumn{3}{c|}{ChrF} & \multicolumn{3}{c|}{COMET} &
 \multicolumn{3}{c}{BLEURT}\\
 \cline{2-13}
~ &All &O &R &All &O &R &All &O &R &All &O &R \\
\hline
Bitext &24.7 & 23.8 & 26.1 & 53.4 & 53.0 & 54.2 & 43.5 & 34.2 & 55.0 & 68.0 & 65.9 & 70.6 \\
\hline
+Beam BT &26.4 & \bf{23.7} & 30.9 & \bf{55.1} & \bf{53.5} & 58.1 & 46.4 & 36.2 & 59.2 & 69.1 & 66.6 & 72.2 \\
\ \ \ \ +TST &\bf{26.6} & 23.5 & \bf{31.8} & 54.9 & 53.1 & \bf{58.4} & \bf{47.8} & \bf{37.4} & \bf{60.5} & \bf{69.5} & \bf{67.0} & \bf{72.7} \\
\hline
\end{tabular}
\caption{Chinese$\rightarrow$English models trained on WMT 2017 Bitext. The Beam BT and the averaged TST results on Original test set (O), Reverse test set (R) and the combined test set (All) from WMT 2017-2019.}
\label{tab:04}
\end{table*}

\begin{table*}
\centering
\setlength{\belowcaptionskip}{-0.5cm}
\begin{tabular}{l|ccc|ccc|ccc|ccc}
\hline
 \multirow{2}* & \multicolumn{3}{c|}{BLEU} & \multicolumn{3}{c|}{ChrF} & \multicolumn{3}{c|}{COMET} &
 \multicolumn{3}{c}{BLEURT}\\
 \cline{2-13}
~ &All &O &R &All &O &R &All &O &R &All &O &R \\
\hline
Bitext &28.7 & 28.8 & 28.6 & 56.0 & 54.1 & 57.9 & 52.5 & 28.8 & 76.3 & 71.6 & 64.7 & 78.5 \\
\hline
+Beam BT &\bf{32.3} & \bf{29.0} & 35.8 & \bf{59.0} & \bf{54.8} & \bf{63.5} & 63.5 & 38.9 & 88.1 & 74.0 & 66.9 & 81.0 \\
\ \ \ \ +TST$_{EnDe}$ & 31.7 & 27.8 & 35.6 & 58.6 & 54.1 & 63.3 & \bf{66.9} & \bf{43.1} & \bf{90.7} & \bf{75.2} & \bf{68.2} & \bf{82.1} \\
\ \ \ \ +TST$_{EnRo}$  &31.9 & 27.8 & \bf{36.1} & 58.6 & 54.0 & \bf{63.5} & 65.0 & 39.9 & 90.2 & 74.5 & 66.9 & \bf{82.1} \\
\hline
\end{tabular}
\caption{English$\rightarrow$Romanian models trained on WMT 2016 bitext (Bitext). Beam BT and TST results on each Original test set (O), Reverse test set (R) and the combined test set (All) from WMT 2016.}
\label{tab:05}
\end{table*}
\subsubsection{Ablation Experiment}
Although TST BT does not directly use additional source text but the transfer model is trained with source data. So we perform forward translation (FT) or self-training \cite{zhang-zong-2016-exploiting} with the same data and compare the FT, FT+BT \cite{wu-etal-2019-exploiting}, and FT + TST BT strategies, as shown in Table \ref{tab:03}. 

FT enhancement is considerable on the original test set but slight on the reverse test set. FT + BT brings significant improvement on the reverse and original test sets. When we perform TST BT on such a strong baseline, we observe further 0.7 and 1.2 COMET score increases on original and reverse sets respectively.

Although FT and TST use the same data, their mechanisms are different and the two methods can be used together. \citet{He2020Revisiting} believe dropout is the key for FT while TST BT focuses on style transfer to create $Nature$ to $Nature$ data, which further improves the translation of $Nature$ input. %从而改善nature输入的翻译质量

\subsection{TST BT for ZhEn}
The size of ZhEn bilingual data is 20M, four times that of EnDe. We perform TST on this language pair to see whether TST BT is effective when applied to a even larger data size and to a language from a different family. We use 20M English monolingual data to ensure the ratio of bilingual and monolingual data is 1:1. See overall results in Table \ref{tab:04} and detailed results in Appendix \ref{app:b}.

The overall result is similar to that of EnDE. We observe significant increase of COMET and BLEURT scores after applying TST BT, although the BLEU and ChrF scores fluctuate. TST BT achieves 1.4 COMET score increase on average on basis of Beam BT. We observe significant increase on both original and reverse test sets.

Our experiments also show that TST BT achieves similar improvements against other BT baselines in addition to Beam BT on ZhEn. The result is different from the EnDe experiment, where the improvement brought by TST against Beam BT is much greater than other BT baselines. We assume that a larger bilingual data size and a different data ratio may be the reason. 
% 这并不像在EnDe上，使用TST之后，Beam BT的提升幅度远远大于其他的BT基线。

It should be noted that the ZhEn baseline is already very strong considering the data size, and even stronger after adding the standard BT data. However, TST BT achieves further enhancement against such strong baselines.
\subsubsection{Human Evaluation}
We also perform human evaluation on ZhEn to verify the enhancement brought by TST BT. We randomly sample 300 sentences from WMT17-19 original and reverse test sets respectively. We follow the evaluation scheme mentioned by \citet{ChrisCallisonBurch2007MetaEO}, and 8 professional annotators are recruited to rate adequacy and fluency of three MT results on a 5-point scale, given source text and reference. 

The result is listed in Table \ref{tab:06}.
% 无论是正向的还是反向测试集，在忠实度和流畅度上，TST都有显著的提升。这个结果与COMET和BLEURT的结果一致。人工评测的结果，再一次验证了我们方法的有效性。而且无论是人工评测，还是自动指标，都能够发现提升的主要方向是Original测试集，说明TST BT对Nature的输入内容起到了改善的作用。--新增
TST improves adequacy and fluency on both original and reverse test sets. The result is consistent with COMET and BLEURT scores in Table \ref{tab:04}. 
The human evaluation result again proves the effectiveness of our method. Both automatic metrics and human evaluations demonstrate that TST BT mainly brings enhancement on the original test set, indicating that TST BT improves the translation of $Nature$ input.
\subsection{TST BT for EnRo}
We further perform experiments in low-resource scenario to test the generalizability of TST BT. We use WMT16 EnRo bilingual data (0.6M bilingual) for the experiment. Table \ref{tab:05} presents the results.

In this experiment, we compare the effectiveness of two TST models: one is trained with EnRo models, and the other, used for our EnDe experiment, is trained with EnDe models. The style transfer model trained with EnRo data improves performance against BT baselines (by 1.5 COMET score and 0.5 BLEURT score). %Such improvement can be measured by COMET and BLEURT. 
%BT baseline performs quite well in the low-resource scenario.

Another interesting finding is that the TST model for the EnDe task also enhances the EnRo model performance (by 3.4 COMET score and 1.2 BLUERT score), which is even greater than that of the TST$_{EnRo}$ model. The result indicates that it is possible to build a universal pre-trained model for sytle transfer.
% 这个点可以再扩展一下：就是说我们的这个发现，除了体现风格迁移的通用性之外，还可能会训练出一个更为通用的预训练模型的可能性--wmh  - xiaoyu done
This result demonstrates that the style transfer model is universal and can be applied to other language pairs. 
%The gap between the two style transfer models may lie in the quality of data used to train the transfer model, since performing RTT on a low-resource language may brought about more errors. We will detail the relationship between BT enhancement and style transfer performance in sex XX. 
\begin{table}
\centering
\setlength{\tabcolsep}{1.25mm}{
\resizebox{\linewidth}{!}{
\begin{tabular}{l|cc|cc}
\hline
 & \multicolumn{2}{c|}{Original} & \multicolumn{2}{c}{Reverse}\\
 \cline{2-5}
 & Adequacy & Fluency & Adequacy & Fluency \\
\hline
Bitext & 3.89 & 4.52 & 4.57 & 4.81 \\ 
\hline
+Beam BT & 3.98 & 4.51 & 4.67 & 4.79 \\
\ \ \ \ +TST & \bf{4.03} & \bf{4.52} & \bf{4.69} & \bf{4.82} \\
\hline
\end{tabular}}}
\caption{Averaged Adequacy and Fluency results by human annotators on Original and Reverse test sets.}
\label{tab:06}
\end{table}

\begin{table}
\centering
\setlength{\belowcaptionskip}{-0.6cm}
\setlength{\tabcolsep}{1.25mm}{
\resizebox{\linewidth}{!}{
\begin{tabular}{l|cccc}
\hline
~ & BLEU & ChrF & COMET & BLEURT \\
\hline
Bitext & 26.1 & 57.1 & 50.4 & 71.0 \\ 
\hline
+Beam BT$_{Med}$ & 28.6 & 60.9  & 56.4 & 72.6 \\
\ \ \ \ +TST$_{Med}$ & \bf{30.3} & \bf{61.0}& \bf{57.8} & \bf{73.3} \\
\hline
\end{tabular}}}
\caption{Metric scores of German$\rightarrow$English models trained on WMT 2018 Bitext. Biomedical Beam BT (Beam BT$_{Med}$) and the biomedical TST (TST$_{Med}$) results measured on WMT 2018 biomedical test set.}
\label{tab:domain}
\end{table}

\subsection{Domain Augmentation}
We observe that the translation of in-domain natural inputs improve significantly after applying TST BT. %To analyze whether it works on out-of-domain natural inputs as well, we measured our EnDe model, which is trained on news data, on Flores and IWSLT 14 test sets (70\% of the two test sets are nature2ht data). The result is shown in Table X. We find that TST BT improves the translation of both in-domain and out-of-domain natural inputs, as BLEU/COMET scores on Flores and IWSLT test sets increase. Particularly, we observe significant improvement on IWSLT test set as BLEU increased by XX and BLEURT by XX. 
% 我们还发现，TST BT对领域外的nature输入依然有显著的提升作用(details see Appendix Table x.x)。如果我们把style换成domain
We also found that TST BT still improve translation of out-of-domain natural inputs (like IWSLT14 and Flores \cite{10.1162/tacl_a_00474}) test set (for details, see Appendix Table \ref{tab:19}). 

%We assume TST BT can also be applied to domain adaptation as along as it can transfer the text to an in-domain style.
Domain adaptation is a critical application of BT. BT can improve in-domain performance given in-domain monolingual data and an out-of-domain translation model \cite{edunov2018understanding}. If we train a TST model to modify the source-side text generated by BT to an in-domain style, we assume in-domain translation can be further improved.

To test our hypothesis, we train an out-of-domain DeEn model using WMT18 news bilingual data, and perform BT on 12M biomedical English monolingual data. 2.5M biomedical German monolingual data is used to train the in-domain TST model. The result is shown in Table \ref{tab:domain}.

We observe significant improvement brought by BT and more surprisingly, further significant improvement after we apply TST, with an increase of 1.4 COMET score and 0.7 BLEURT score. We believe the reason for such enhancement is the same as that on Flores and IWSLT test sets mentioned above: making the input style biased towards in-domain or $Nature$ text to augment the effectiveness of BT. The experiment again demonstrates the generalizability of TST BT.%和BT方法一样，依然能够在领域增强中起作用

\begin{figure}
\centering
\setlength{\belowcaptionskip}{-0.4cm}
\includegraphics[height=5.5cm,width=7.5cm]{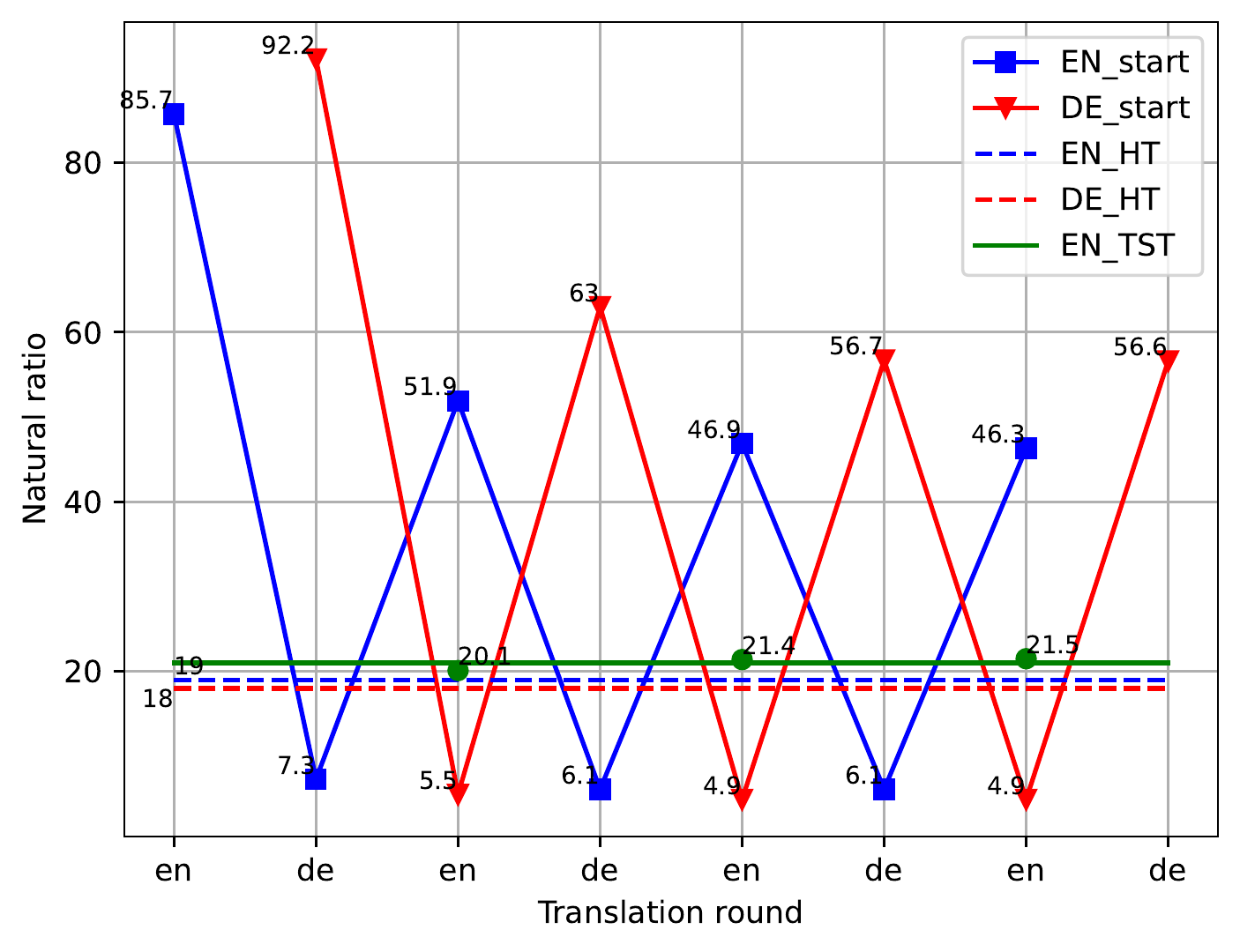}
\caption{The $Nature$ ratio of each round of translation results starting with $EN_{Nature}$ and $DE_{Nature}$ (EN\_start, DE\_start). The dotted line indicates the $Nature$ ratio of English or German human translations (EN\_HT, DE\_HT). The green line represents the averaged $Nature$ ratio (EN\_TST) of English data after style transfer.}
\label{figure:04}
% 英文单语和德语数据经过多轮的翻译，每轮翻译结果的nature程度占比。虚线表示英语或者德语人工翻译结果的nature程度。绿线表示nature程度很低的英语，经过TST之后的Nature程度均值。
\end{figure}

\section{Analysis}
\subsection{Style Tide}
\label{sec:6.1}
As shown in Figure \ref{figure:DataTrain}, bilingual data can be divided into $Nature$ to $HT$ or $HT$ to $Nature$. By learning such data, the model inclines to generate translation-like output when the input is $Nature$, and vice versa. To illustrate the phenomenon, we perform several rounds of translation on $EN_{Nature}$ and $DE_{Nature}$ data from WMT18 EnDe test set. We calculate the proportion of $Nature$ text marked by the classifier after each round of translation. 

As shown in Figure \ref{figure:04}, the percentage of $Nature$ sentences fluctuates regularly after each round of translation, no matter the translation starts from De or En. For English original data, the percentage of $Nature$ data is 85.7\% before translation. The percentage drops to 7.3\% after the first round of translation into German, and then bounces back to 51.9 after the second round of translation back into English. As we analyzed above, the style of input determines the style of output. 
%In general $Nature$ input produces HT-like outputs and vice versa. 
%RTT only slightly reduces the proportion of Nature text, and the overall Nature percentage is much higher than HT. 

 In general, the composition of bilingual data, as well as the difference between $Nature$ and $HT$ style, makes the source-side BT text significantly different from $Nature$ text. As a result, the translation of $Nature$ input can hardly be improved by standard BT. 

 % 我们还发现，因为风格潮汐的现象，通过RTT制造的英文，导致英文本身的nature程度很高，远远高于德语翻译出来的英语(91.5 vs 7.3)。通过RTT方法制造的TST训练数据，
 
 %However, transferring source-side BT text to a more natural style is a zero-shot issue and RTT may not be able to complete this task. 
\subsection{Style and Quality of TST}
%TST only modifies the source-side text translated by a BT model. 
To understand what changes a TST model makes to the source-side text, we analyze the style and quality difference before and after applying TST to the source-side text. 

% 以EnDe为例，我们分析TST前后两种英语数据的风格变化，并且通过人工评测，对比TST前后两种数据的质量变化。
% 如Figure \ref{fig:04}所示，TST之后风格更偏向于Nature了，从xx提升到xx。提升还是非常显著的，基本持平HT的nature程度，但是和单语的Nature程度还是有一定的差距。
Taking EnDe data as an example, we analyze the style of English text before and after TST, and compare the quality through human evaluation.

As shown in Figure \ref{figure:04}, after TST, the percentage of $Nature$ text increases from 5.5 to 20.1. The improvement is significant, reaching the same level of $Nature$ as human-translated data, but there is still a certain gap with the real natural text.
%In other words, both human translations and TST BT data are significantly different from real natural text in terms of style. However, TST BT greatly enhances the translation of natural inputs despite the style difference. 

In addition, to analyze the impact of TST on text quality, we randomly select 300 sentences from WMT14 test set and assess the quality of standard BT data and TST data against references. We invite three professional annotators to complete the assessment. We use relative ranking and classify the results into three categories: equal, TST better or MT better. The result is shown in Table \ref{tab:RR}, which is different from \citet{freitag2019ape}. APE can further improve translation quality but TST cannot.

Based on above analysis, we find that TST does not improve the overall quality of source-side BT data. Instead, it modifies the text towards a more natural style, thus overcomes the weakness of standard BT. In addition, TST BT still maintains BT's tolerance \cite{NikolayBogoychev2019DomainTA} of data quality to make up the performance deterioration caused by TST. 

\subsection{Style Transfer and BT Effects}
\label{sec:6.3}
%这个表格估计还是要的
%\begin{table}
%\centering
%\scalebox{0.85}{
%\begin{tabular}{l|ccccc|c}
%\hline
%\textbf{} & 2014 & 2015 & 2016 & 2017 & 2018 & avg\\
%\hline
%HT &13.0 & 19.2 & 16.5 & 20.5 & 16.1 & 17.1 \\
%\hline
%MT &4.1 & 4.2 & 3.8 & 5.6 & 5.5 & 4.6 \\
%\ \ \ \ +TST &\bf{21.7} & \bf{21.3} & \bf{20.2} & \bf{26.6} & \bf{20.1} & \bf{22.0} \\
%\hline
%\hline
%Nature &92.1 & 85.4 & 90.4 & 87.5 & 85.7 & 88.2 \\
%\hline
%\end{tabular}}
%四种类型的英语数据：Original测试集的英语（Nature），Reverse测试集的英语Human Translation（HT）和Reverse测试集的德语翻译成英语的结果（MT），以及通过TST转换的英语，他们被分类器标识为nature的比例。
%\caption{The percentage of data classified as $nature$ in different types of data.}
%\label{tab:5}
%\end{table}

\begin{table}
\centering
\setlength{\belowcaptionskip}{-0.4cm}
\begin{tabular}{l|ccc}
\hline
\textbf{} & Equal  & TST better  & MT better  \\
\hline
Annotator 1 &220 &27 & 53   \\
\hline
Annotator 2 &212 &23  &65   \\
\hline
Annotator 3 &218 &24 & 58 \\
\hline
\end{tabular}
\caption{Evaluation results by professional annotators on 300 randomly selected MT and TST sentences.}
\label{tab:RR}
\end{table}

\begin{table}
\centering
\setlength{\belowcaptionskip}{-0.5cm}

\begin{tabular}{l|cc|c}
\hline
 & \multicolumn{2}{c|}{\textbf{TST}} & \textbf{TST BT} \\
\cline{2-4}
 & ACC & Seman & COMET \\
 \hline
EnDe Beam BT & 5.5 & 71.1 & 35.3 \\ 
\hline
+TST$_{Direct}$ & \bf{25.2} & \bf{70.8} & 51.1 \\
+TST$_{Cascade}$ & 20.1 & 69.9 & \bf{59.3} \\
\hline
\hline
EnRo Beam BT & 25.4 & 65.6 & 38.9 \\ 
\hline
+TST$_{EnDe}$ & \bf{55.4} & \bf{65.1} & \bf{43.1} \\
+TST$_{EnRo}$ & 52.2 & 65.7 & 39.9 \\
\hline
\end{tabular}
\caption{Style transfer performances of different TST models on WMT 2018 English$\rightarrow$German and WMT 2016 English$\rightarrow$Romanian translation tasks.}
\label{tab:TSTBT}
\end{table}

% 为了进一步分析风格迁移的效果和BT最终的提升效果之间的关系，我们对比EnDe和EnRo中使用使用的两个风格迁移器的效果。
% 我们用测试集英文中被分类器分成Nature的数据比例来表示风格迁移的力度（ACC）, 用迁移后英语和参考译文的BLEURT值表示语义的保留程度（Seman）。 
% 以EnDe为例，我们用EnDe的反向测试集的De来做BT并通过参考译文来计算语义保留程度。
% BT最终的提升效果，我们用正向测试集的COMET结果来衡量。
% 结果如表x所示，EnRo上风格迁移准确率更好的TST，最终在BT增强后的模型机器翻译质量提升的更多一些。但是在EnD上，这种抢情况就变的有点异常，CTST方法无论是在迁移的准确度还是语义的保留程度都要好于TST，但是最终BT的提升效果却弱于TST。
% 风格迁移的程度和BT提升效果的关系，我们会在未来的工作中进一步的分析。
In order to analyze the relationship between style transfer results and final improvement on translation quality, we compare the improvements brought by TST BT data that is generated via two different approaches (cascaded/direct as we motioned above) on EnDe and EnRo. 

%We use the proportion of data in the test set that is classified as Nature by the classifier to indicate the strength of style transfer (ACC), and use the BLEURT value of the transferred English and reference translations to indicate the degree of semantic retention (Seman).

We use Strength of Style Transfer (ACC) and Semantic Preservation (Seman) to measure style transfer results. Taking EnDe as an example, we perform BT on the $DE_{nature}$ data from the reverse test set $\{EN_{HT}$, $DE_{nature}\}$, and calculate Seman (measured by BLEURT) against reference $EN_{HT}$. We then use the original test set $\{EN_{Nature}$, $DE_{HT}\}$ to measure the improvement of TST BT on the translation of $Nature$ input. The result shows that although the direct approach leads to higher ACC and Seman scores, the cascaded approach brings greater enhancement to the final translation performance. The results are shown in Table \ref{tab:TSTBT}. 

For EnRo, we compare style transfer models trained on EnRo 
%(TST$_{EnRo}$) 
and EnDe 
%(TST$_{EnDe}$)
data as we stated before. Data modified by the TST$_{EnDe}$ achieves higher ACC and Seman scores, and lead to greater enhancement to the overall translation quality. The result is different from our EnDe experiment.

Therefore, the relationship between style transfer and the effect of BT enhancement can not be drawn and more researches are required. 

\section{Conclusion}
This paper proposes Text Style Transfer Back Translation (TST BT) to further enhance BT effectiveness. We make a detailed analysis of training data styles and find that BT hardly improves translation of $Natural$ inputs, which are the main inputs in practical use. Our method simply modifies the style of source-side BT data, which brings significant improvements on translation quality, both in high-resource and low-resource language scenarios. Further experiment finds that TST BT is also effective in domain adaptation, which can further expand the application of our method. The generalizability of TST BT is thus proved. 
\section{Limitations}
TST BT is simple and straightforward, which brings great improvements against BT baselines. However, comparing with standard BT, TST BT requires an additional style transfer model and additional time to process generated BT data. 

\bibliography{anthology,custom}
% \bibliographystyle{acl_natbib}
% \bibliography{acl_natbib}

\appendix
\onecolumn
\section{Experiment Details for EnDe}
\label{EnDeDetails}
% \FloatBarrier
% \vspace{-80\baselineskip}

\begin{table}[htbp]
\centering
\setlength{\tabcolsep}{1.1mm}{\begin{adjustbox}{width=\columnwidth,center}
\begin{tabular}{l|ccc|ccc|ccc|ccc|ccc}
\hline
 \multirow{2}* & \multicolumn{3}{c|}{2014} & \multicolumn{3}{c|}{2015} & \multicolumn{3}{c|}{2016} &
 \multicolumn{3}{c|}{2017} & 
 \multicolumn{3}{c}{2018}\\
 \cline{2-16}
~  &All &O &R &All &O &R &All &O &R &All &O &R &All &O &R\\
\hline
Bitext &28.2 & 28.2& 28.3& 30.8& 32.8& 26.1& 34.6& 37.6& 29.8& 28.7& 31.1& 25.2& 42.2& 46.3& 34.9 \\
\hline
+Beam BT &28.8 & 23.7& 35.2& 28.9& 27.6& 30.6& 33.2& 28.7& 39.3& 29.2& 26.1& 32.3& 40.2& 36.2& 44.5 \\
\ \ \ \ +Direct-TST & 30.1 & 26.5 & 34.2 & 30.2 & 29.9 & 29.8 & 34.7 & 32.5 & 37.1 & 29.4 & 27.8 & 30.4 & 42.1 & 40.5 & 42.6\\
\ \ \ \ +Cascade-TST &\bf{32.1} & \bf{28.5}& \bf{36.7}& \bf{32.3}& \bf{31.9}& \bf{32.5}& \bf{36.4}& \bf{33.7}& \bf{40.0}& \bf{31.5}& \bf{28.9}& \bf{33.8}& \bf{44.0}& \bf{42.0} & \bf{45.4} \\
\hline
+Sampling BT & 33.7 & 28.8& \bf{39.4}& \bf{33.8}& \bf{32.4}& 35.7& \bf{36.5} & \bf{32.9} & \bf{41.7}& \bf{32.0}& \bf{28.3} & \bf{36.4}& \bf{44.0} & 40.9& \bf{47.8} \\
\ \ \ \ +TST  &\bf{33.8} & \bf{29.3}& 38.9& 33.5& 32.0 & \bf{35.9} & 36.4& 32.6& \bf{41.7}& 31.7& 28.1& 36.0& 43.8& \bf{41.0} & 47.1 \\
\hline
+Noised BT &32.5 & 29.8& 36.1& 33.4& 34.3& 31.5& 38.4& 38.0 & 38.4& 31.9& 31.5& 31.8& \bf{46.6}& \bf{47.2}& 44.1 \\
\ \ \ \ +TST  &\bf{33.0} & \bf{30.4}& \bf{36.4}& \bf{34.2}& \bf{34.9}& \bf{32.8}& \bf{38.8}& \bf{38.1}& \bf{39.4}& \bf{32.6}& \bf{32.0}& \bf{32.7}& 46.5& 47.1& \bf{44.3} \\
\hline
+Tagged BT &32.7 & 30.0& 36.0& 34.1& 34.4& \bf{32.3}& 38.7& 38.6& \bf{38.7}& \bf{32.9}& 32.6& \bf{32.3}& 46.8& 47.6& \bf{44.4}\\
\ \ \ \ +TST  &\bf{33.0} & \bf{30.6}& \bf{36.1}& \bf{34.5}& \bf{35.6}& 31.8& \bf{39.3}& \bf{39.8}& 38.3& \bf{32.9}& \bf{32.7}& 32.2& \bf{47.3}& \bf{48.5}& 44.2\\
\hline
\hline
+FT &28.7 & 29.2& 28.1& 31.5& 33.8& 26.1& 35.6& 38.8& 30.3& 29.5& 32.5& 25.3& 42.9& 47.7& 34.9 \\
\ \ \ \ +Beam BT  &32.3 & \bf{30.1}& 35.2& 33.7& \bf{35.1}& 30.5& 39.8& 40.4& 38.5& 32.7& 32.6& 32.0& 47.9& 49.6& 44.2 \\
\ \ \ \ \ \ \ \ +TST  &\bf{32.7} & 30.0& \bf{36.0}& \bf{34.2}& \bf{35.1}& \bf{31.9}& \bf{40.5}& \bf{40.5}& \bf{40.0}& \bf{33.4}& \bf{33.2}& \bf{33.0}& \bf{48.3}& \bf{49.7}& \bf{45.0} \\
\hline
\end{tabular}\end{adjustbox}}
\caption{English$\rightarrow$German BLEU scores on WMT 2014-2018 test sets.}
\label{tab:10}
\end{table}

\begin{table}[htbp]
\centering
\setlength{\tabcolsep}{1.1mm}{\begin{adjustbox}{width=\columnwidth,center}
\begin{tabular}{l|ccc|ccc|ccc|ccc|ccc}
\hline
 \multirow{2}* & \multicolumn{3}{c|}{2014} & \multicolumn{3}{c|}{2015} & \multicolumn{3}{c|}{2016} &
 \multicolumn{3}{c|}{2017} & 
 \multicolumn{3}{c}{2018}\\
 \cline{2-16}
~  &All &O &R &All &O &R &All &O &R &All &O &R &All &O &R\\
\hline
Bitext &58.7 & 58.4 & 59.0 & 58.9 & 60.0 & 56.7 & 61.9 & 63.3 & 60.2 & 57.8 & 59.2 & 56.2 & 66.8 & 69.6 & 63.2 \\
\hline
+Beam BT &57.4 & 51.8 & 64.9 & 56.4 & 54.1 & 61.2 & 60.1 & 54.9 & 66.9 & 57.4 & 53.8 & 61.9 & 64.6 & 60.4 & 70.1 \\
\ \ \ \ +Direct-TST & 59.2  & 55.4  & 64.3  & 58.3  & 57.2  & 60.7  & 61.8  & 58.9  & 65.7  & 58.0  & 56.0  & 60.6  & 66.6  & 64.7  & 69.0\\
\ \ \ \ +Cascade-TST &\bf{61.4} & \bf{58.2} & \bf{65.8} & \bf{60.3} & \bf{59.3} & \bf{62.4} & \bf{63.7} & \bf{60.9} & \bf{67.5} & \bf{60.1} & \bf{58.1} & \bf{62.7} & \bf{68.4} & \bf{66.6} & \bf{70.8} \\
\hline
+Sampling BT &61.9 & 58.3 & \bf{66.9} & \bf{61.0} & \bf{59.7} & \bf{63.7} & 63.6 & \bf{60.3} & 68.1 & \bf{60.3} & \bf{57.4} & \bf{63.9} & \bf{68.1} & 65.4 & \bf{71.8} \\
\ \ \ \ +TST  &\bf{62.1} & \bf{58.8} & 66.7 & 60.8 & 59.4 & \bf{63.7} & \bf{63.7} & \bf{60.3} & \bf{68.4} & 60.2 & \bf{57.4} & 63.8 & \bf{68.1} & \bf{65.6} & 71.4 \\
\hline
+Noised BT &62.0 & 59.5 & 65.3 & 61.0 & 60.8 & 61.5 & 64.9 & 63.7 & 66.5 & 60.3 & 59.3 & 61.6 & 69.9 & 69.8 & 70.1 \\
\ \ \ \ +TST  &\bf{62.4} & \bf{60.1} & \bf{65.6} & \bf{61.8} & \bf{61.5} & \bf{62.4} & \bf{65.2} & \bf{63.9} & \bf{67.0} & \bf{61.1} & \bf{60.1} & \bf{62.2} & \bf{70.0} & \bf{69.9} & \bf{70.2} \\
\hline
+Tagged BT &62.0 & 59.8 & 65.0 & 61.2 & 61.0 & \bf{61.7} & 65.1 & 64.1 & \bf{66.3} & 60.9 & 60.4 & 61.6 & 70.2 & 70.4 & \bf{70.0}\\
\ \ \ \ +TST  &\bf{62.3} & \bf{60.3} & \bf{65.1} & \bf{61.7} & \bf{61.8} & 61.5 & \bf{65.6} & \bf{65.1} & \bf{66.3} & \bf{61.2} & \bf{60.8} & \bf{61.7} & \bf{70.5} & \bf{70.9} & 69.9\\
\hline
\hline
+FT &59.3 & 59.3 & 59.4 & 59.5 & 60.9 & 56.6 & 62.7 & 64.2 & 60.5 & 58.6 & 60.2 & 56.5 & 67.6 & 70.9 & 63.3 \\
\ \ \ \ +Beam BT  &62.0 & \bf{59.9} & 65.0 & 61.4 & 61.5 & 61.2 & 65.9 & 65.5 & 66.4 & 61.0 & 60.4 & 61.7 & 71.0 & 71.7 & 70.0 \\
\ \ \ \ \ \ \ \ +TST  &\bf{62.2} & 59.8 & \bf{65.4} & \bf{61.7} & \bf{61.7} & \bf{61.9} & \bf{66.4} & \bf{65.6} & \bf{67.5} & \bf{61.5} & \bf{60.9} & \bf{62.3} & \bf{71.3} & \bf{71.9} & \bf{70.4} \\
\hline
\end{tabular}\end{adjustbox}}
\caption{English$\rightarrow$German ChrF scores on WMT 2014-2018 test sets.}
\label{tab:11}
\end{table}

\begin{table}[htbp]
\centering
\setlength{\tabcolsep}{1.1mm}{\begin{adjustbox}{width=\columnwidth,center}
\begin{tabular}{l|ccc|ccc|ccc|ccc|ccc}
\hline
 \multirow{2}* & \multicolumn{3}{c|}{2014} & \multicolumn{3}{c|}{2015} & \multicolumn{3}{c|}{2016} &
 \multicolumn{3}{c|}{2017} & 
 \multicolumn{3}{c}{2018}\\
 \cline{2-16}
~  &All &O &R &All &O &R &All &O &R &All &O &R &All &O &R\\
\hline
Bitext &54.4 & 51.2 & 57.6 & 51.4 & 49.8 & 54.4 & 54.5 & 46.2 & 62.7 & 51.9 & 44.7 & 59.1 & 61.8 & 58.7 & 64.9 \\
\hline
+Beam BT &44.3 & 23.8 & 64.8 & 41.0 & 32.3 & 57.9 & 46.0 & 22.5 & 69.5 & 44.8 & 25.9 & 63.7 & 53.3 & 35.3 & 71.3 \\
\ \ \ \ +Direct-TST & 52.4 & 39.6 & 65.2 & 49.4 & 44.6 & 58.6 & 53.1 & 37.8 & 68.3 & 50.4 & 37.8 & 63.1 & 61.4 & 51.1 & 71.8\\
\ \ \ \ +Cascade-TST &\bf{59.0} & \bf{52.0} & \bf{66.1} & \bf{55.9} & \bf{52.6} & \bf{62.5} & \bf{59.1} & \bf{47.4} & \bf{70.8} & \bf{56.3} & \bf{46.8} & \bf{65.7} & \bf{66.1} & \bf{59.3} & \bf{72.9} \\
\hline
+Sampling BT &61.6 & 55.0 & \bf{68.4} & 58.7 & 55.8 & 64.4 & 61.5 & 51.0 & 72.0 & 59.0 & 50.0 & \bf{68.0} & 67.6 & 60.7 & \bf{74.6} \\
\ \ \ \ +TST  &\bf{62.7} & \bf{57.3} & 68.0 & \bf{59.5} & \bf{56.6} & \bf{65.1} & \bf{62.3} & \bf{52.1} & \bf{72.4} & \bf{59.7} & \bf{51.4} & \bf{68.0} & \bf{68.3} & \bf{62.2} & 74.5 \\
\hline
+Noised BT &60.1 & 54.5 & 65.6 & 55.7 & 53.0 & 61.1 & 60.1 & 50.2 & 70.0 & 56.1 & 47.1 & 65.1 & 67.2 & 62.0 & 72.5 \\
\ \ \ \ +TST  &\bf{62.0} & \bf{57.6} & \bf{66.4} & \bf{57.8} & \bf{55.7} & \bf{62.1} & \bf{62.8} & \bf{54.2} & \bf{71.4} & \bf{59.7} & \bf{53.0} & \bf{66.3} & \bf{69.1} & \bf{64.8} & \bf{73.3} \\
\hline
+Tagged BT &61.4 & 56.2 & \bf{66.7} & 57.8 & 55.6 & \bf{62.0} & 61.7 & 52.9 & \bf{70.6} & 58.6 & 51.6 & 65.7 & 68.3 & 63.9 & 72.8\\
\ \ \ \ +TST  &\bf{61.8} & \bf{57.4} & 66.3 & \bf{58.2} & \bf{56.2} & \bf{62.0} & \bf{62.4} & \bf{54.2} & \bf{70.6} & \bf{59.3} & \bf{52.6} & \bf{66.0} & \bf{69.2} & \bf{65.4} & \bf{73.1}\\
\hline
\hline
+FT &56.5 & 53.5 & 59.6 & 52.8 & 51.9 & 54.6 & 55.7 & 48.3 & 63.0 & 53.2 & 47.0 & 59.4 & 63.3 & 61.3 & 65.4 \\
\ \ \ \ +Beam BT  &60.6 & \bf{55.7} & 65.5 & 56.7 & 54.7 & 60.5 & 60.1 & 51.0 & 69.2 & 57.2 & 49.2 & 65.2 & 67.3 & 63.0 & 71.5 \\
\ \ \ \ \ \ \ \ +TST  &\bf{60.8} & 55.4 & \bf{66.1} & \bf{57.6} & \bf{55.5} & \bf{61.8} & \bf{61.1} & \bf{51.5} & \bf{70.7} & \bf{58.3} & \bf{50.4} & \bf{66.2} & \bf{68.7} & \bf{64.2} & \bf{73.2} \\
\hline
\end{tabular}\end{adjustbox}}
\caption{English$\rightarrow$German COMET scores on WMT 2014-2018 test sets.}
\label{tab:12}
\end{table}

\begin{table}[htbp]
\centering
\setlength{\tabcolsep}{1.1mm}{\begin{adjustbox}{width=\columnwidth,center}
\begin{tabular}{l|ccc|ccc|ccc|ccc|ccc}
\hline
 \multirow{2}* & \multicolumn{3}{c|}{2014} & \multicolumn{3}{c|}{2015} & \multicolumn{3}{c|}{2016} &
 \multicolumn{3}{c|}{2017} & 
 \multicolumn{3}{c}{2018}\\
 \cline{2-16}
~  &All &O &R &All &O &R &All &O &R &All &O &R &All &O &R\\
\hline
Bitext &73.5 & 71.9 & 75.2 & 72.7 & 72.2 & 73.7 & 73.5 & 70.5 & 76.5 & 72.7 & 69.8 & 75.6 & 75.8 & 74.5 & 77.2 \\
\hline
+Beam BT &71.5 & 65.3 & 77.7 & 69.9 & 67.3 & 75.1 & 72.2 & 65.2 & 79.1 & 71.3 & 65.0 & 77.5 & 73.8 & 68.3 & 79.4 \\
\ \ \ \ +Direct-TST & 73.8 & 69.6 & 78.0 & 72.5 & 71.2 & 75.1 & 74.0 & 69.3 & 78.7 & 73.0 & 68.6 & 77.3 & 76.3 & 72.9 & 79.7\\
\ \ \ \ +Cascade-TST &\bf{75.7} & \bf{72.9} & \bf{78.5} & \bf{74.5} & \bf{73.5} & \bf{76.5} & \bf{75.9} & \bf{72.1} & \bf{79.7} & \bf{74.9} & \bf{71.4} & \bf{78.5} & \bf{77.8} & \bf{75.4} & \bf{80.3} \\
\hline
+Sampling BT &77.1 & 74.4 & \bf{79.7} & 75.7 & 74.7 & 77.6 & 76.8 & 73.1 & 80.5 & 76.0 & 72.7 & 79.3 & 78.7 & 76.1 & \bf{81.4} \\
\ \ \ \ +TST  &\bf{77.4} & \bf{75.1} & \bf{79.7} & \bf{76.0} & \bf{75.2} & \bf{77.8} & \bf{77.2} & \bf{73.7} & \bf{80.7} & \bf{76.3} & \bf{73.1} & \bf{79.5} & \bf{78.9} & \bf{76.5} & 81.3 \\
\hline
+Noised BT &75.9 & 73.4 & 78.4 & 74.2 & 73.1 & 76.3 & 75.8 & 72.2 & 79.4 & 74.6 & 70.9 & 78.2 & 77.8 & 75.5 & 80.1 \\
\ \ \ \ +TST  &\bf{76.5} & \bf{74.3} & \bf{78.7} & \bf{75.0} & \bf{74.2} & \bf{76.6} & \bf{76.8} & \bf{73.6} & \bf{80.1} & \bf{75.7} & \bf{72.9} & \bf{78.6} & \bf{78.7} & \bf{76.8} & \bf{80.6} \\
\hline
+Tagged BT &76.2 & 73.7 & 78.6 & 74.8 & 74.0 & 76.3 & 76.3 & 72.9 & \bf{79.7} & 75.3 & 72.2 & 78.3 & 78.4 & 76.5 & 80.3\\
\ \ \ \ +TST  &\bf{76.5} & \bf{74.3} & \bf{78.8} & \bf{75.2} & \bf{74.5} & \bf{76.6} & \bf{76.7} & \bf{73.7} &\bf{ 79.7} & \bf{75.6} & \bf{72.6} & \bf{78.6} & \bf{78.8} & \bf{76.9} & \bf{80.6}\\
\hline
\hline
+FT &74.3 & 72.8 & 75.8 & 73.0 & 72.7 & 73.6 & 73.9 & 71.3 & 76.5 & 73.1 & 70.5 & 75.7 & 76.2 & 75.3 & 77.2 \\
\ \ \ \ +Beam BT  &75.8 & 73.5 & 78.1 & 74.2 & 73.5 & 75.4 & 75.5 & 72.1 & 79.0 & 74.6 & 71.2 & 77.9 & 77.9 & 76.0 & 79.7 \\
\ \ \ \ \ \ \ \ +TST  &\bf{76.2} & \bf{73.8} & \bf{78.6} & \bf{74.8} & \bf{74.0} & \bf{76.2} & \bf{76.1} & \bf{72.5} & \bf{79.6} & \bf{75.1} & \bf{71.9} & \bf{78.4} & \bf{78.5} & \bf{76.6} & \bf{80.3} \\
\hline
\end{tabular}\end{adjustbox}}
\caption{English$\rightarrow$German BLEURT scores on WMT 2014-2018 test sets.}
\label{tab:13}
\end{table}

\FloatBarrier
\vspace{-1000\baselineskip}

\section{Experiment Details for ZhEn}
\label{app:b}

\begin{table}[htbp]
\centering
\tiny
\setlength{\tabcolsep}{1.1mm}{
\begin{adjustbox}{width=\columnwidth,center}
\begin{tabular}{l|ccc|ccc|ccc|ccc}
\hline
 \multirow{2}* & \multicolumn{3}{c|}{2017} & \multicolumn{3}{c|}{2018} & \multicolumn{3}{c|}{2019} &
 \multicolumn{3}{c}{Average} \\
 \cline{2-13}
~  &All &O &R &All &O &R &All &O &R &All &O &R \\
\hline
Bitext & 24.4 & 24.5 & 24.2 & 24.8 & 23.0 & 28.3 & 24.8 & 24.0 & 25.7 & 24.7 & 23.8 & 26.1 \\
\hline
+Beam BT & 25.7 & 23.2 & 29.1 & 26.5 & 23.3 & 32.8 & \bf{27.1} & \bf{24.5} & 30.8 & 26.4 & \bf{23.7} & 30.9\\
\ \ \ \ +TST & \bf{26.1} & \bf{23.3} & \bf{30.1} & \bf{26.9} & \bf{23.6} & \bf{33.8} & 26.7 & 23.7 & \bf{31.5} & \bf{26.6} & 23.5 & \bf{31.8}\\
\hline
+Sampling BT & \bf{26.2} & \bf{22.7} & \bf{31.2} & 26.6 & 22.9 & \bf{34.1} & \bf{26.9} & \bf{23.5} & \bf{32.3} & \bf{26.6} & 23.0 & \bf{32.5} \\
\ \ \ \ +TST & \bf{26.2} & 22.6 & 31.1 & \bf{26.8} & \bf{23.4} & 33.9 & 26.8 & \bf{23.5} & 32.2 & \bf{26.6} & \bf{23.2} & 32.4 \\
\hline
+Noised BT & \bf{26.3} & \bf{24.4} & \bf{28.9} & 26.6 & 23.7 & \bf{32.5} & \bf{27.0} & 24.7 & \bf{30.6} & 26.6 & 24.3 & \bf{30.7}\\
\ \ \ \ +TST & 26.1 & 24.2 & 28.6 & \bf{26.9} & \bf{24.1} & \bf{32.5} & \bf{27.0} & \bf{24.8} & 30.5 & \bf{26.7} & \bf{24.4} & 30.5\\
\hline
+Tagged BT & \bf{26.3} & \bf{24.3} & \bf{29.0} & 26.6 & 23.7 & 32.5 & \bf{27.1} & \bf{24.5} & \bf{31.0} & \bf{26.7} & \bf{24.2} & \bf{30.8}\\
\ \ \ \ +TST & 25.7 & 23.6 & 28.7 & \bf{27.0} & \bf{24.2} & \bf{32.7} & 27.0 & \bf{24.5} & \bf{31.0} & 26.6 & 24.1 & \bf{30.8}\\
\hline
\end{tabular}
\end{adjustbox}}
\caption{Chinese$\rightarrow$English BLEU scores on WMT 2017-2019 test sets.}
\label{tab:14}
\end{table}

\begin{table}[htbp]
\centering
\tiny
\setlength{\tabcolsep}{1.1mm}{
\begin{adjustbox}{width=\columnwidth,center}
\begin{tabular}{l|ccc|ccc|ccc|ccc}
\hline
 \multirow{2}* & \multicolumn{3}{c|}{2017} & \multicolumn{3}{c|}{2018} & \multicolumn{3}{c|}{2019} &
 \multicolumn{3}{c}{Average} \\
 \cline{2-13}
~  &All &O &R &All &O &R &All &O &R &All &O &R \\
\hline
Bitext & 53.4  & 54.0  & 52.5  & 53.2  & 52.0  & 55.7  & 53.5  & 53.1  & 54.3  & 53.4  & 53.0  & 54.2\\
\hline
 +Beam BT & \bf{54.7}  & \bf{53.6}  & 56.2  & 54.8  & \bf{52.6}  & 59.7  & \bf{55.8}  & \bf{54.3}  & 58.5  & \bf{55.1}  & \bf{53.5}  & 58.1\\
\ \ \ \  +TST & 54.3  & 52.9  & \bf{56.4}  & \bf{54.9}  & \bf{52.6}  & \bf{60.0}  & 55.6  & 53.8  & \bf{58.7}  & 54.9  & 53.1  & \bf{58.4} \\
\hline
 +Sampling BT & \bf{54.4}  & \bf{52.4}  & \bf{57.2}  & \bf{54.5}  & 52.0  & \bf{60.2}  & \bf{55.0}  & \bf{52.7}  & \bf{59.0}  & \bf{54.6}  & \bf{52.4}  & \bf{58.8} \\
\ \ \ \ +TST & 54.1  & 51.9  & 57.1  & \bf{54.5}  & \bf{52.1}  & 59.9  & 54.8  & \bf{52.7}  & 58.7  & 54.5  & 52.2  & 58.6 \\
\hline
 +Noised BT & \bf{54.7}  & \bf{53.9}  & \bf{55.7}  & 54.6  & 52.7  & \bf{58.9}  & \bf{55.3}  & \bf{53.8}  & \bf{58.1}  & \bf{54.9}  & \bf{53.5}  & \bf{57.6} \\
\ \ \ \  +TST & 54.4  & 53.6  & 55.5  & \bf{54.7}  & \bf{52.9}  & \bf{58.9}  & 55.1  & 53.7  & 57.8  & 54.7  & 53.4  & 57.4 \\
\hline
 +Tagged BT & \bf{54.7}  & \bf{53.9}  & \bf{55.8}  & 54.5  & 52.5  & \bf{59.0}  & \bf{55.2}  & \bf{53.6}  & \bf{58.2}  & \bf{54.8}  & \bf{53.3}  & \bf{57.7}\\
\ \ \ \  +TST & 54.3  & 53.3  & 55.7  & \bf{54.8 } & \bf{52.9}  & \bf{59.0}  & 55.1  & 53.4  & 58.1  & 54.7  & 53.2  & 57.6\\
\hline
\end{tabular}
\end{adjustbox}}
\caption{Chinese$\rightarrow$English ChrF scores on WMT 2017-2019 test sets.}
\label{tab:15}
\end{table}

\begin{table}[htbp]
\centering
\setlength{\abovecaptionskip}{0pt}
\setlength{\belowcaptionskip}{0pt}
\tiny
\setlength{\tabcolsep}{1.1mm}{
\begin{adjustbox}{width=\columnwidth,center}
\begin{tabular}{l|ccc|ccc|ccc|ccc}
\hline
 \multirow{2}* & \multicolumn{3}{c|}{2017} & \multicolumn{3}{c|}{2018} & \multicolumn{3}{c|}{2019} &
 \multicolumn{3}{c}{Average} \\
 \cline{2-13}
~  &All &O &R &All &O &R &All &O &R &All &O &R \\
\hline
Bitext & 46.6  & 39.7  & 53.4  & 39.5  & 29.2  & 56.4  & 44.4  & 33.7  & 55.1  & 43.5  & 34.2  & 55.0\\
\hline
 +Beam BT & 48.9  & 40.0  & 57.8  & 42.4  & 31.0  & 61.3  & 48.0  & 37.6  & 58.4  & 46.4  & 36.2  & 59.2 \\
\ \ \ \  +TST & \bf{49.7}  & \bf{40.3}  & \bf{59.0}  & \bf{44.3}  & \bf{33.1}  & \bf{62.9}  & \bf{49.3}  & \bf{38.8}  & \bf{59.7}  & \bf{47.8}  & \bf{37.4}  & \bf{60.5} \\
\hline
 +Sampling BT & 48.6  & 37.0  & 60.1  & 42.4  & 30.4  & 62.3  & 47.8  & 34.7  & \bf{60.9}  & 46.3  & 34.0  & 61.1 \\
\ \ \ \  +TST & \bf{49.9}  & \bf{39.2}  & \bf{60.5}  & \bf{43.5}  & \bf{31.8}  & \bf{62.9}  & \bf{48.6}  & \bf{36.5}  & 60.6  & \bf{47.3}  & \bf{35.8}  & \bf{61.3}  \\
\hline
 +Noised BT & 49.7  & 41.1  & \bf{58.4}  & 43.3  & 32.3  & \bf{61.5}  &\bf{ 48.8}  & 37.6  & \bf{60.1}  & 47.3  & 37.0  & \bf{60.0} \\
\ \ \ \  +TST & \bf{50.1}  & \bf{41.9}  & 58.2  &\bf{ 43.8}  & \bf{33.3}  & 61.1  & 48.7  & \bf{38.1}  & 59.8  & \bf{47.5}  & \bf{37.8}  & 59.7\\
\hline
 +Tagged BT & \bf{49.7}  & \bf{41.1}  & 58.2  & 43.1  & 31.9  & \bf{61.7}  & 48.3  & 36.4  & \bf{60.2}  & 47.0  & 36.5  & \bf{60.0} \\
\ \ \ \  +TST & 49.3  & 40.1  & \bf{58.4}  & \bf{44.1}  & \bf{33.8}  & 61.4  & \bf{48.8}  & \bf{37.7}  & 60.0  & \bf{47.4}  & \bf{37.2}  & 59.9\\
\hline
\end{tabular}
\end{adjustbox}}
\caption{Chinese$\rightarrow$English COMET scores on WMT 2017-2019 test sets.}
\label{tab:16}
\end{table}

\begin{table}[htbp]
\centering
\tiny
\setlength{\tabcolsep}{1.1mm}{
\begin{adjustbox}{width=\columnwidth,center}
\begin{tabular}{l|ccc|ccc|ccc|ccc}
\hline
 \multirow{2}* & \multicolumn{3}{c|}{2017} & \multicolumn{3}{c|}{2018} & \multicolumn{3}{c|}{2019} &
 \multicolumn{3}{c}{Average} \\
 \cline{2-13}
~  &All &O &R &All &O &R &All &O &R &All &O &R \\
\hline
Bitext& 67.7 & 65.9 & 53.4 & 67.6 & 65.2 & 71.6 & 68.8 & 66.7 & 70.8 & 68.0 & 65.9 & 70.6 \\
\hline
 +Beam BT& 68.6 & 66.2 & 57.8 & 68.7 & 65.9 & 73.3 & 70.0 & 67.7 & 72.3 & 69.1 & 66.6 & 72.2\\
\ \ \ \  +TST& \bf{68.9} & \bf{66.3} & \bf{59.0} & \bf{69.3} & \bf{66.6} & \bf{73.8} & \bf{70.4} & \bf{68.0} & \bf{72.8} & \bf{69.5} & \bf{67.0} & \bf{72.7}\\
\hline
 +Sampling BT& 68.7 & 65.5 & 60.1 & 68.7 & 65.5 & 73.9 & 70.0 & 67.0 & 73.1 & 69.1 & 66.0 & 73.0 \\
\ \ \ \  +TST& \bf{69.0} & \bf{66.1} & \bf{60.5} & \bf{69.1} & \bf{66.1} & \bf{74.0} & \bf{70.3} & \bf{67.3} & \bf{73.3} & \bf{69.5} & \bf{66.5} & \bf{73.1} \\
\hline
 +Noised BT& \bf{68.9} & \bf{66.5} & \bf{58.4} & 68.9 & 66.2 & \bf{73.3} & 70.1 & 67.6 & \bf{72.7} & \bf{69}.3 & 66.8 & \bf{72.4}\\
\ \ \ \  +TST& 68.8 & \bf{66.5} & 58.2 & \bf{69.0} & \bf{66.4} & \bf{73.3} & \bf{70.2} & \bf{67.8} & 72.6 & \bf{69.3} & \bf{66.9} & 72.3\\
\hline
 +Tagged BT& \bf{68.9} & \bf{66.6} & 58.2 & 68.8 & 66.0 & \bf{73.5} & 70.1 & 67.4 & \bf{72.8} & 69.3 & 66.7 & \bf{72.5} \\
\ \ \ \  +TST& 68.7 & 66.2 & \bf{58.4} & \bf{69.1} &\bf{66.5} & \bf{73.5} & \bf{70.3} & \bf{67.8} & \bf{72.8} & \bf{69.4} & \bf{66.8} & \bf{72.5} \\
\hline
\end{tabular}
\end{adjustbox}}
\caption{Chinese$\rightarrow$English BLEURT scores on WMT 2017-2019 test sets.}
\label{tab:17}
\end{table}

\FloatBarrier
\vspace{-1000\baselineskip}

\section{Experiment Details for EnRo}

\begin{table}[htbp]
\centering
\begin{adjustbox}{width=\columnwidth,center}
\begin{tabular}{l|ccc|ccc|ccc|ccc}
\hline
 \multirow{2}* & \multicolumn{3}{c|}{BLEU} & \multicolumn{3}{c|}{ChrF} & \multicolumn{3}{c|}{COMET} &
 \multicolumn{3}{c}{BLEURT}\\
 \cline{2-13}
~ &All &O &R &All &O &R &All &O &R &All &O &R \\
\hline
Bitext &28.7 & 28.8 & 28.6 & 56.0 & 54.1 & 57.9 & 52.5 & 28.8 & 76.3 & 71.6 & 64.7 & 78.5 \\
\hline
+Beam BT &\bf{32.3} & \bf{29.0} & 35.8 & \bf{59.0} & \bf{54.8} & \bf{63.5} & 63.5 & 38.9 & 88.1 & 74.0 & 66.9 & 81.0 \\
\ \ \ \ +TST$_{EnDe}$ & 31.7 & 27.8 & 35.6 & 58.6 & 54.1 & 63.3 & \bf{66.9} & \bf{43.1} & \bf{90.7} & \bf{75.2} & \bf{68.2} & \bf{82.1} \\
\ \ \ \ +TST$_{EnRo}$  &31.9 & 27.8 & \bf{36.1} & 58.6 & 54.0 & \bf{63.5} & 65.0 & 39.9 & 90.2 & 74.5 & 66.9 & \bf{82.1} \\
\hline
+Sampling BT &\bf{32.6} & \bf{29.3} & \bf{35.9} & \bf{59.0} & \bf{54.8} & \bf{63.5} & 66.0 & 42.1 & 90.1 & 75.1 & 68.0 & 82.2 \\
\ \ \ \ +TST$_{EnDe}$ &31.9 & 28.2 & 35.7 & 58.5 & 54.1 & 63.2 & \bf{66.9} & \bf{42.7} & \bf{91.2} & \bf{75.4} & \bf{68.2} & \bf{82.5} \\
\ \ \ \ +TST$_{EnRo}$ &32.1 & 28.4 & \bf{35.9} & 58.5 & 54.2 & 63.1 & 65.4 & 40.8 & 90.0 & 75.2 & 67.9 & \bf{82.5} \\
\hline
+Noised BT &32.2 & 30.8 & \bf{33.7} & 58.8 & 55.9 & \bf{62.0} & 66.8 & 45.0 & 88.5 & 75.3 & 68.7 & 81.9 \\
\ \ \ \ +TST$_{EnDe}$ &32.3 & 31.6 & 33.1 & \bf{59.2} & \bf{56.8} & 61.7 & \bf{68.7} & \bf{47.5} & \bf{90.0} & \bf{76.2} & \bf{70.0} & \bf{82.4} \\
\ \ \ \ +TST$_{EnRo}$ &\bf{32.7} & \bf{31.9} & \bf{33.7} & 59.1 & 56.5 & 61.8 & 67.3 & 45.9 & 88.7 & 75.7 & 69.1 & \bf{82.4} \\
\hline
+Tagged BT & 32.5 & 31.7 & 33.2 & 58.9 & 56.5 & 61.5 & 67.7 & 45.8 & 89.6 & 75.7 & 69.1 & 82.2 \\
\ \ \ \ +TST$_{EnDe}$ &\bf{33.1} & \bf{32.6} & \bf{33.7} & \bf{59.3} & \bf{57.1} & \bf{61.7} & \bf{70.3} & \bf{49.8} & \bf{90.7} & \bf{76.8} & \bf{70.9} & \bf{82.8} \\
\ \ \ \ +TST$_{EnRo}$ &32.6 & 32.0 & 33.3 & 59.0 & 56.7 & 61.5 & 68.2 & 46.5 & 89.8 & 76.1 & 69.5 & 82.7 \\
\hline
\end{tabular}\end{adjustbox}
\caption{WMT 2016 English$\rightarrow$Romanian results on the WMT 2016 test set.}
\label{tab:18}
\end{table}

% \vspace{-1000\baselineskip}
% \FloatBarrier
% \vspace{-1000\baselineskip}

\section{TST BT on OOD}

% \vspace{-100\baselineskip}
\begin{table}[htbp]
\centering
% \setlength{\abovecaptionskip}{0pt}
% \setlength{\belowcaptionskip}{0pt}
% \huge
\setlength{\tabcolsep}{1.2mm}{
\begin{adjustbox}{width=\columnwidth,center}
{\begin{tabular}{l|cccc|cccc|cccc}
\hline
 \multirow{2}* & \multicolumn{4}{c|}{Flores} & \multicolumn{4}{c|}{IWSLT-2014} & \multicolumn{4}{c}{Med-2021} \\
 \cline{2-13}
~ & BLEU & ChrF & COMET & BLEURT & BLEU & ChrF & COMET & BLEURT & BLEU & ChrF & COMET & BLEURT \\
\hline
Bitext & 34.5 & 61.8 & 55.3 & 74.2 & 28.5 & 55.8 & 34.8 & 68.7 & 23.5 & 54.8 & 46.8 & 71.1 \\
\hline
Beam BT & 33.7 & 60.9 & 53.2 & 73.5 & 24.5 & 50.6 & 18.5 & 64.6 & 25.9 & \bf{57.6} & 49.4 & 71.7  \\
+TST & \bf{34.9} & \bf{62.2} & \bf{59.6} & \bf{75.8} & \bf{27.9} & \bf{55.2} & \bf{38.0} & \bf{69.7} & \bf{26.3} & 57.0 & \bf{49.9} & \bf{72.2}  \\
\hline
Tagged BT & 37.2 & 63.6 & 61.1 & 76.0 & 29.7 & 56.7 & 40.4 & 70.0 & 25.5 & 56.1 & 49.1 & 71.9 \\
+TST & \bf{38.2} & \bf{64.0} &\bf{61.8} & \bf{76.4} & \bf{30.0} & \bf{57.1} & \bf{41.3} & \bf{70.4} & \bf{25.8} & \bf{56.6} & \bf{50.6} & \bf{72.3} \\
\hline
\end{tabular}}\end{adjustbox}}
\caption{WMT 2018 English$\rightarrow$German results on out-of-domain (Flores, IWSLT and Medical) original test set.}
\label{tab:19}
\end{table}

\end{document}